\documentclass[10pt,journal,compsoc]{IEEEtran}
\usepackage{amsmath,amssymb,amsfonts}
\usepackage{booktabs}
\usepackage{times}
\usepackage{footnote}
\usepackage{multirow}
\usepackage{subcaption}
\usepackage{textcomp}
\usepackage{stfloats}
\usepackage{url}
\usepackage{verbatim}
\usepackage{graphicx}
\usepackage{wrapfig}
\usepackage{csquotes}
\usepackage{xcolor, colortbl}
\usepackage{mathrsfs}
\usepackage{algorithm}
\usepackage{algpseudocode}
\usepackage{arydshln}
\usepackage{listings}
\definecolor{maroon}{cmyk}{0,0.87,0.68,0.32}

\definecolor{Gray}{gray}{0.9}
\definecolor{LightCyan}{rgb}{0.85,1,1}

\ifCLASSOPTIONcompsoc
  \usepackage[nocompress]{cite}
\else
  \usepackage{cite}
\fi

\ifCLASSINFOpdf
\else

\fi

\begin{document}

\title{From Missing Pieces to Masterpieces: Image Completion with Context-Adaptive Diffusion}

\author{Pourya~Shamsolmoali,~\IEEEmembership{Senior Member,~IEEE,}
        Masoumeh~Zareapoor,~\IEEEmembership{Member,~IEEE,} Huiyu~Zhou,
        Michael~Felsberg,~\IEEEmembership{Senior Member,~IEEE},
        Dacheng~Tao,~\IEEEmembership{Fellow,~IEEE},
        and Xuelong~Li,~\IEEEmembership{Fellow,~IEEE}

\thanks{P.~Shamsolmoali is with the Department of Computer Science, University of York, UK and SCEE, East China Normal University, China (pshams55@gmail.com)}
\thanks{M.~Zareapoor (corresponding author) is with the Department of Electrical Engineering, Shanghai Jiao Tong University, China (mzarea222@gmail.com).}
\thanks{H.~Zhou is with the School of Computing and Mathematical Sciences, University of Leicester, UK (hz143@leicester.ac.uk).}
\thanks{M.~Felsberg is with Computer Vision Laboratory, Linkoping University, Sweden (michael.felsberg@liu.se).}
\thanks{D.~Tao is with the College of Computing and Data Science at Nanyang Technological University, Singapore (dacheng.tao@gmail.com).}
\thanks{X.~Li is with the Institute of Artificial Intelligence (TeleAI) of China Telecom (xuelong\_li@ieee.org).}
\thanks{P. Shamsolmoali and M. Zareapoor contributed equally.}
}


\IEEEtitleabstractindextext{%
\begin{abstract}
Image completion is a challenging task, particularly when ensuring that generated content seamlessly integrates with existing parts of an image. While recent diffusion models have shown promise, they often struggle with maintaining coherence between known and unknown (missing) regions. This issue arises from the lack of explicit spatial and semantic alignment during the diffusion process, resulting in content that does not smoothly integrate with the original image. Additionally, diffusion models typically rely on global learned distributions rather than localized features, leading to inconsistencies between the generated and existing image parts. In this work, we propose ConFill, a novel framework that introduces a Context-Adaptive Discrepancy (CAD) model to ensure that intermediate distributions of known and unknown regions are closely aligned throughout the diffusion process. By incorporating CAD, our model progressively reduces discrepancies between generated and original images at each diffusion step, leading to contextually aligned completion. Moreover, ConFill uses a new Dynamic Sampling mechanism that adaptively increases the sampling rate in regions with high reconstruction complexity. This approach enables precise adjustments, enhancing detail and integration in restored areas. Extensive experiments demonstrate that ConFill outperforms current methods, setting a new benchmark in image completion.
\end{abstract}

}

\maketitle

\IEEEdisplaynontitleabstractindextext

\IEEEpeerreviewmaketitle

\ifCLASSOPTIONcompsoc
\IEEEraisesectionheading{\section{Introduction}\label{sec:introduction}}
\else
\section{Introduction}
\label{sec:introduction}
\fi

\IEEEPARstart{I}{mage} completion, also called image inpainting, is an important task in computer vision and image processing that aims to reconstruct missing or corrupted portions of an image such that the restored areas are both visually pleasing and coherent. This technique is widely used in various domains, including image editing \cite{ling2021editgan, li2022sdm}, image restoration \cite{he2019adversarial, li2022all}, and the removal of unwanted objects \cite{ yildirim2023inst}.
At present, several state-of-the-art methods use approaches such as GANs \cite{suvorov2022resolution, lu2022glama, zeng2022aggregated}, transformers \cite{li2022mat, shamsolmoali2023transinpaint}, and autoregressive modeling \cite{wan2021high, shamsolmoali2023vtae}. Despite their advancements, these techniques often experience unstable training \cite{shamsolmoali2023vtae} and struggle with filling large missing regions in images \cite{zeng2022aggregated, li2022sdm}.
Recently, there has been a growing interest among researchers in using diffusion models \cite{ho2020denoising, song2020denoising} to address the above issue. These generative models transform noisy images into natural-looking images through sequential denoising steps. A common strategy in this domain is to use a pre-trained, static diffusion model, which eliminates the need for additional training and enhances the model's flexibility and adaptability.
\begin{figure}
\centering
\includegraphics[width=0.98\columnwidth]{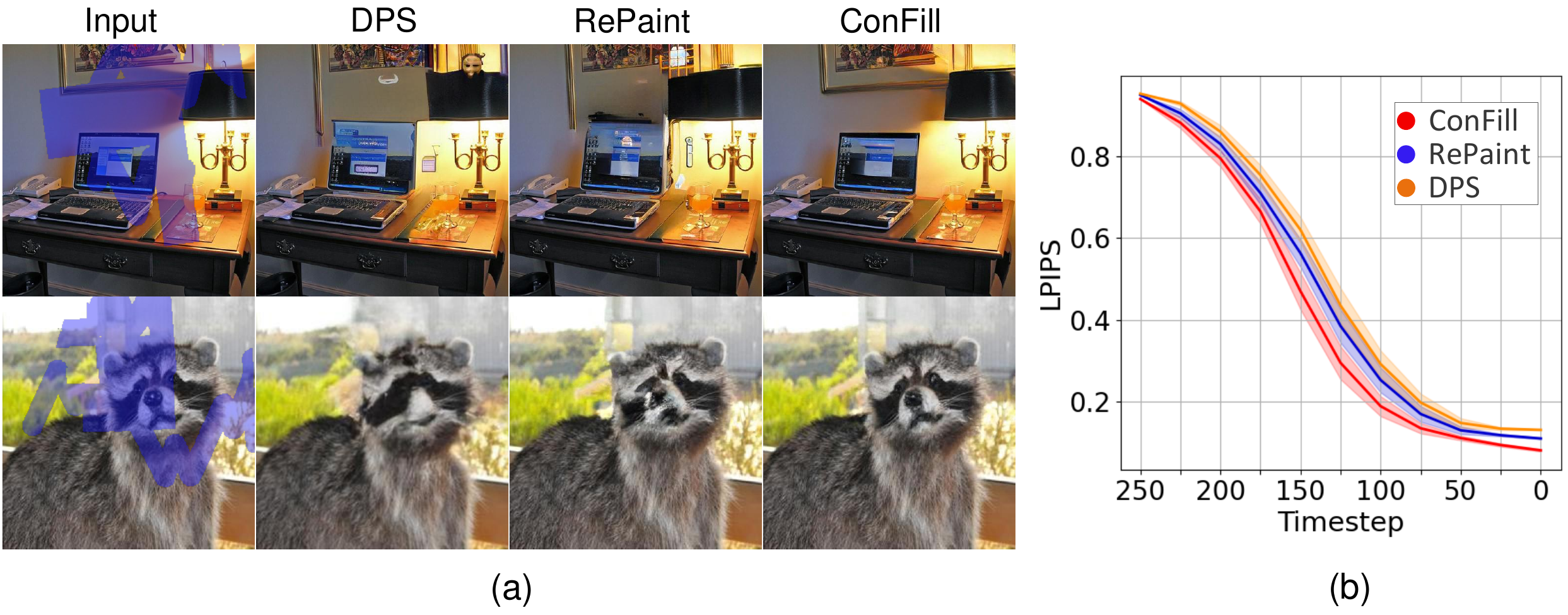} 
\vspace{-6pt}
\caption {\small{(a) Generated images by DPS, RePaint, and our ConFill, using a fixed diffusion model on different masked inputs. (b) LPIPS score vs. denoising timesteps for a single input image.}}
\label{fig1}
\end{figure}
These methods address image completion constraints by using various replacement strategies. For example, by replacing predicted image regions with corresponding parts from the reference image \cite{wang2022zero}, or by incorporating corrupted regions from the reference image into the intermediate denoising process \cite{avrahami2022blended, lugmayr2022repaint}. Similarly, \cite{li2022sdm} focuses on optimizing pixel distribution rather than denoising, aiming to reduce the number of iterations needed for training.
Although these methods modify large portions of an image, they struggle to properly adjust the pixels in unknown areas to ensure consistency with the surrounding context, resulting in noticeable discontinuities, which significantly compromise the overall quality of the generated images \cite{trippe2022diffusion}. Additionally, the sampling techniques used in \cite{kong2021fast, jonnarth2022importance, liu2022pseudo} often compromise sample quality by introducing additional noise, further limiting their effectiveness.
%
In image completion, the efficacy of diffusion models is significantly influenced by the discrepancy metrics they use \cite{tong2021diffusion, li2023dpm}. These metrics are important as they determine the fidelity of the transition from noisy image to an output that aligns with the original image regions. Traditional discrepancy metrics used in diffusion models, such as Euclidean distance, generally focus on minimizing statistical disparities without considering the complex spatial and structural details of the images. For example, two images might have similar pixel values but different textures or structures, leading to high Euclidean distance but visually incoherent outputs \cite{ozair2019wasserstein}.
Moreover, the limitations of traditional discrepancy metrics become more visible when models are required to adapt to the specific needs of different regions of an image. Indeed, each part of an image may have unique characteristics such as texture, lighting, or context that require specific handling to achieve a realistic completion. DPS \cite{chung2022diffusion} and SMCDiff \cite{trippe2022diffusion} use Bayesian optimization for conditioned sampling, however, their reliance on computationally intensive processes and approximations compromises both accuracy and visual quality. Fig. \ref{fig1} illustrates examples of incoherent image completions for masked images. DPS shows a noticeable disparity between the left and right sides of the generated images. RePaint also has discontinuities in the reconstructions. 
MCG \cite{chung2022improving} uses manifold constraints, but this can restrict diffusion process, leading to underfitting and failing to capture complex data variability. 

CoPaint \cite{zhang2023towards} addresses this problem by modifying both the known and unknown regions to achieve more seamless integration. This approach alters key image features, such as edges that should remain intact, leading to errors that compromise the objectives of image completion.
In these diffusion models, the focus is on learning the overall noise distribution, with limited attention to the alignment between consecutive denoising steps. While this assumption of treating each step as independent can be effective for completing small missing regions, it becomes problematic when dealing with large or complex areas. Without explicitly modeling the transitions between steps, inconsistencies can arise, leading to artifacts or unrealistic completions, particularly in regions requiring detailed reconstruction. 

A potential solution is to optimize the mapping between inverse diffusion steps. Recently, \cite{li2023dpm} uses Brenier theorem \cite{lu2019brenier, paty2020regularity} to ensure structural fidelity during transformations between two probability distributions for generative task. Brenier theorem provides a globally optimal solution for mapping distributions, but it does not consider local coherence within patches of the image \cite{papadakis2015optimal}. In diffusion-based image completion, maintaining smooth and coherent transitions across these local patches is essential. This gap highlights the need for a more flexible alignment process that can better handle the diverse patterns and textures within an image. In this paper, we introduce ConFill, a novel framework specially designed for image completion. ConFill overcomes the limitations of traditional discrepancy metrics in diffusion models and enhances both precision and contextual relevance in image reconstruction. At the core of ConFill is the Context-adaptive Discrepancy (CAD) method, which redefines the denoising process by minimizing inter-step discrepancies in the diffusion sequence, ensuring smoother transitions and improving alignment with the original image content. CAD incorporates a context-sensitive adaptation mechanism that dynamically adjusts to local image attributes throughout the diffusion process. This ensures that each image region is processed based on its unique characteristics, thereby enhancing the alignment of the reconstructed image with the original content. As a result, every step in the diffusion cycle is optimized for better contextual consistency.
Moreover, ConFill enhances image reconstruction quality by selectively sampling from the posterior distribution of intermediate images, which helps reduce errors and improve the quality of the denoised output. 

Indeed, ConFill through combination of CAD and dynamic sampling bridges the gap between local and global features, dynamically aligning the diffusion process with both the immediate context and the broader structure of the image.
Additionally, the framework optimises the iterative process of final image generation, leading to more efficient computation and enhancing the overall performance of the model.

%
In our experimental analysis, ConFill is evaluated on three challenging datasets using masks of varying shapes and sizes. The results demonstrate that ConFill outperforms existing diffusion-based methods in achieving superior fidelity and coherence in image completion tasks. These findings are supported by enhanced performance metrics in both qualitative and quantitative analyses, highlighting the model's effectiveness in complex image completion. 
Specifically, ConFill achieves a relative improvement of $10.9\%$ and $6.7\%$ in LPIPS scores compared to the leading baselines, RePaint and CoPaint, respectively, on the ImageNet dataset. Our main contributions are as follows:

\begin{itemize}
\item We introduce ConFill, a unified learning framework that integrates our CAD model with an enhanced diffusion model, providing a robust solution for image completion. This integration enables the model to adaptively refine image completions by using contextual features, resulting in higher precision in denoising and improved coherence across the completed images.
\item ConFill uses principles from the Brenier potential to determine the transformation that minimizes the discrepancy between latent distributions across different timesteps. This approach reduces mode mixing, resulting in high-fidelity samples that closely match the target data distribution and effectively address the inverse problem.
\item We introduce a Dynamic Sampling mechanism that adjusts sampling density based on structural complexity. This strategy improves the focus and efficiency of the diffusion process, particularly in regions with higher complexity.

\end{itemize} 

The rest of this manuscript is structured as follows: Section 2 discusses related studies. In Section 3, we introduce the details of ConFill. Section 4 discusses the experimental results and ablation study. Finally, Section 5 provides the concluding remarks.

\section{Related Work}
\subsection{Image Completion}
Conventional image completion models are based on robust low-level assumptions, such as using local patches \cite{barnes2009patchmatch} to reconstruct missing regions in images. Recent advancements in CNN-based methods for image inpainting \cite{iizuka2017globally, yu2018generative, zhang2018semantic, liu2020rethinking, zeng2021cr, suvorov2022resolution, dong2022incremental, lu2022glama, shamsolmoali2023vtae, jain2023keys} generally adopt encoder-decoder structures or variations of these stacked architectures. In particular, \cite{iizuka2017globally} introduces a framework based on GANs that uses both local and global discriminators. CoMod-GAN \cite{zhao2021large} is another remarkable method that enhances imagine generation quality by including a stochastic noise vector in the encoded representation. Building upon the GAN framework, innovative components have been integrated, including regional convolutions \cite{liu2018image}, regional normalization \cite{yu2020region} and attentions \cite{yu2018generative, zeng2021cr}. New advances such as the Vision Transformer \cite{dosovitskiy2020image} are also attracting significant interest in the image inpainting community \cite{wan2021high, yu2021diverse, li2022mat, shamsolmoali2023transinpaint}.
Meaningful priors, including semantic labels \cite{liao2021image} and edge information \cite{nazeri2019edgeconnect}, perform an important role in improving the process. To generate high-quality completions, multi-stage networks are a popular approach. For instance, \cite{yi2020contextual} demonstrates refinement of low-resolution outputs through contextual aggregation. MAT \cite{li2022mat} and ZITS \cite{dong2022incremental} represent transformer-based image inpainting systems specifically designed for high-resolution images. However, their multi-stage network structure tends to be slow for practical, real-world applications. In contrast, LaMa \cite{suvorov2022resolution} introduces a one-stage approach by integrating multi-scale receptive fields to enhance efficiency. However, the speed of these systems is still limited by the conventional decoders they use.

While these approaches have demonstrated promising results in image completion tasks, they largely rely on supervised learning, where networks are trained on specific types of degradation. These approaches demand considerable computational resources and often struggle with masks not seen during training, resulting in limited generalization capabilities \cite{li2022sdm}. More recently, diffusion methods have been gaining popularity for their impressive performance in image generation tasks \cite{bond2022unleashing, chung2022come, bansal2022cold, liu2022delving, su2022dual, wu2023slotdiffusion, zhang2022gddim}. An additional advantage of these methods is their ability to perform image completion tasks effectively without the need for training on specific types of degradation \cite{song2019generative}. In this section, we discuss existing diffusion-based image completion methods, classifying them into two primary categories: supervised and unsupervised approaches \cite{kawar2022denoising}.

\subsection{Diffusion Image Completion}
Supervised diffusion-based image completion methods require training a diffusion model designed for completion tasks, with a focus on various forms of image degradation. In \cite{saharia2022palette}, the diffusion model receives the corrupted image at every stage of its training process. A similar approach is adopted by \cite{nichol2021glide} and \cite{xie2023smartbrush}, in which a text-conditional diffusion model is specifically designed for completion tasks. In \cite{rombach2022high}, an autoencoding model is integrated to compress the image space, followed by the concatenation of spatially aligned conditioning information with the model's input. In contrast, \cite{chung2022come} uses a non-expansive mapping approach to aggregate the degradation operation throughout the training process. In \cite{wang2022zero}, a conditional diffusion model is proposed, designed to focus on refining and improving outputs from a deterministic predictor through specific training of the diffusion model. Additionally, \cite{song2022pseudoinverse} and \cite{liu20232} address a range of inverse problems by directly estimating conditional scores based on the specific measurement model without needing extra training. \cite{chung2022improving} addresses the inverse problems in diffusion models by integrating manifold constraints and \cite{chung2024direct} introduces an enhanced inference method by modeling variable noise levels at the pixel level.
Recent studies have explored extending diffusion models with external guidance. ControlNet \cite{zhang2023adding} integrates structured conditions like edge maps, enabling precise control over outputs. Similarly, PowerPaint \cite{zhuang2025task} uses task-specific prompts to guide inpainting tasks. Both these approaches highlight the adaptability of diffusion models to specific tasks through conditioning.
However, these approaches require training for specific types of degradation, leading to intensive computational demands and limited adaptability to degradation operators not used in the training dataset.

\begin{figure}[t]
\centering
\includegraphics[width=0.95\columnwidth]{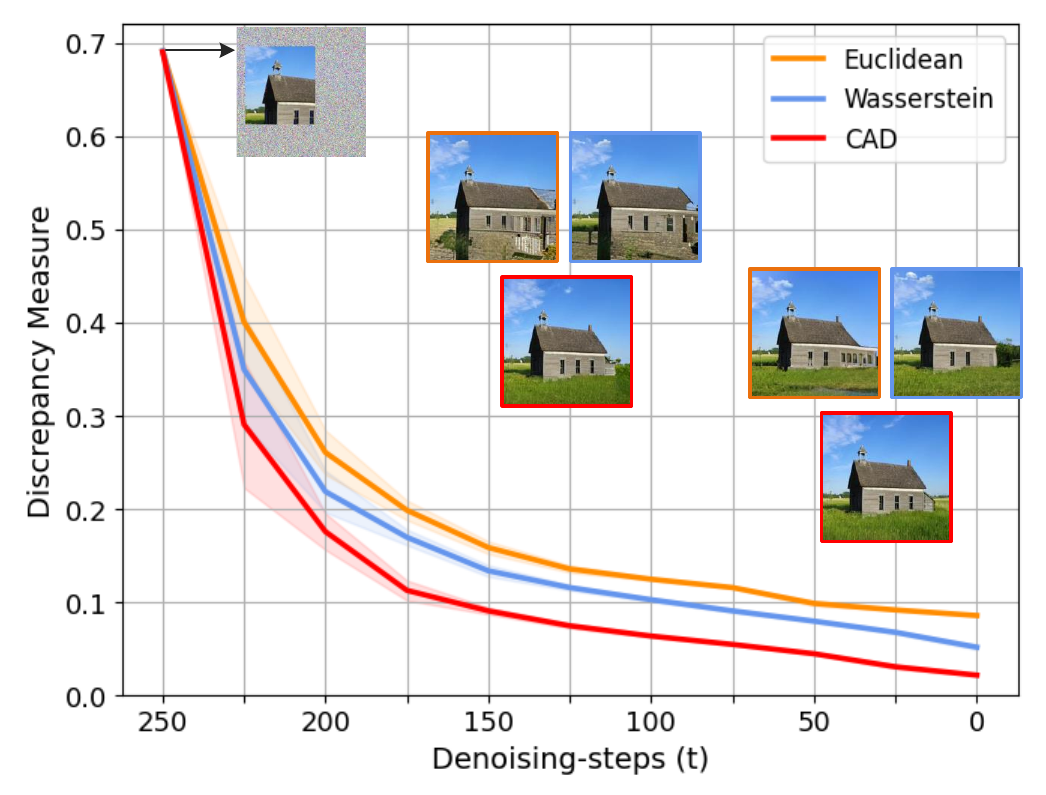} 
\vspace{-6pt}
\caption{\small{The evolution of the discrepancy during the denoising process. This illustrates the mean pixel-wise discrepancy using Context-Adaptive Discrepancy (CAD) (red), Wasserstein Discrepancy (blue), and Euclidean Distance (orange).}}
\label{fig2}
\end{figure} 
Unlike supervised approaches, unsupervised diffusion image completion approaches rely on pre-trained diffusion models, which eliminates the requirement for model structure adjustments. Our method is categorised as unsupervised diffusion. \cite{song2019generative} proposes a modification to the DDPM (denoising diffusion probabilistic models) \cite{ho2020denoising} sampling process, involving the integration of a noisy variant of the damaged image at every stage of the denoising sequence. This approach is similarly applied in \cite{avrahami2022blended} for text-driven image completion. In DDRM \cite{kawar2022denoising}, a posterior diffusion process is formulated, demonstrating that its marginal probability distribution aligns with that of the DDPM framework. This denoising process can be approximated as a weighted-sum blending of the degraded image iteratively at each timestep. Even if these blending-based techniques are effective, the images they generate often lack harmony in the recovered parts, leading to inconsistencies in the generated images \cite{lugmayr2022repaint}. 

To address this challenge, RePaint \cite{lugmayr2022repaint} introduces a new resampling strategy that incorporates a bidirectional operation in diffusion time. This approach blends images from the current timestep \( t \) with a noisy version of the degraded image, enhancing the reconstruction. The resulting blended images are then used to generate outputs for the next timestep \( t+1 \) through a one-step forward process. By incorporating this mechanism, RePaint effectively reduces the visual inconsistencies commonly associated with basic blending methods. In \cite{trippe2022diffusion}, it is demonstrated that methods based on simple blending can lead to irreducible approximation errors during the image generation process.
To address this, a particle filtering-based approach (residual resampling) is proposed. In this method, for each timestep $t$, every generated image is resampled according to its likelihood of producing the known region of the degraded image at the previous timestep $t-1$. This method aims to enhance the accuracy of the generation process by aligning it more closely with the known parts of the degraded image. \cite{pokle2022deep} explores diffusion models through the lens of deep equilibrium analysis, leading to the development of an efficient method for inverting denoising diffusion implicit models \cite{song2020denoising}. This approach focuses on optimising memory usage to make diffusion models more effective. DDNM \cite{wang2022zero} employs a blending strategy that incorporates the damaged image directly into each timestep of the diffusion process without adding any noise. Meanwhile, DPS \cite{chung2022diffusion} addresses the task of completion by adopting a method similar to classifier-free guided diffusion \cite{dhariwal2021diffusion}. This method is focused on approximating posterior sampling, introducing a refined strategy for image completion.
In a similar direction, CoPaint \cite{zhang2023towards} proposes an efficient inpainting framework based on \cite{song2020denoising}. By using Markov diffusion processes, this method enables a deterministic sampling strategy that accelerates inference while maintaining high-quality reconstructions. However, despite its efficiency, the reliance on deterministic sampling limits diversity in plausible completions, especially in complex inpaintings. 
Different from these methods, our approach refines the completion process by progressively aligning the missing regions with the surrounding context through an adaptive refinement strategy. By incorporating structural and textural constraints into the denoising process, our method ensures consistency across timesteps while maintaining high fidelity in complex regions. This allows for more accurate reconstructions that better preserve global coherence and fine-grained details.

\vspace{-3pt}
\section{OUR METHOD}
\subsection{Problem Statement}
Image completion is a process designed to fill in damaged or missing areas of an image \( x_0 \) with content that is contextually relevant and visually coherent. This objective can be written as
\begin{equation}
\begin{array}{lr}
\begin{aligned}
\mathcal{L}_{\mathrm{completion}} = \|s(\hat{x}_0) - r_0\|_2^2 + \digamma \|\hat{x}_0 - P_{\text{data}}\|_2^2,
\end{aligned}
\end{array}
\label{4-1}
\end{equation}
\noindent where \( s(\cdot) \) is an operator that extracts the known parts of the image \( r_0 \), \( \| \cdot \|_2 \) denotes the $L_2$ norm and \(P_{\text{data}}\) represents the natural image prior, while $\digamma$ controls the trade-off between these two objectives. Minimizing this loss ensures that the reconstructed image \( \hat{x}_0 \) aligns with the known portions \( r_0 \).

Diffusion models are widely used as effective methods for image generation, excelling at modeling complex data distributions. These models are particularly well-suited for image completion due to their iterative denoising process, which progressively refines noisy inputs into high-quality images. In image completion, diffusion models start with a noisy version of the image and iteratively refine it to produce a plausible output. In the forward diffusion process, noise is progressively added into the image,
\begin{equation}
\begin{array}{lr}
\begin{aligned}
q(x_{1:T} | x_0) = \prod_{t=1}^{T} p(x_t | x_{t-1}),
\end{aligned}
\end{array}
\label{4-2}
\end{equation}
\noindent in which \( x_t \) denotes the noisy image at timestep \( t \), and \( T \) represents the number of timesteps. Moreover, to control the noise level, we can define the forward diffusion as $q(x_t | x_{t-1}) = \mathcal{N}(x_t; \sqrt{1 - \beta_t} x_{t-1}, \beta_t \text{I})$, in which \(\mathcal{N}\) denotes a Gaussian distribution, \(\beta_t\) denotes the variance of the noise added at timestep \( t \), \(\sqrt{1 - \beta_t}\) scales the previous image \( x_{t-1} \), and \text{I} indicates the identity matrix.
In contrast, the goal of the reverse operation is to recover \( x_0 \) from \( x_T \),
\begin{equation}
\begin{array}{lr}
\begin{aligned}
p_\theta(x_{t-1} | x_t) = \mathcal{N}(x_{t-1}; \mu_\theta(x_t, t), \sigma_\theta(x_t, t)),
\end{aligned}
\end{array}
\label{4-3}
\end{equation}
\begin{algorithm}\small
\caption{CAD}
\label{alg1}
\begin{algorithmic}[1]
\Require Reference image $r_0$, pre-trained denoising functions $g_\theta$, $G$, adaptability function parameters $Var(x)$ and $\psi$
\Ensure Transport map, minimizing CAD

\State Initialize transport map $\mathbf{f}: \vartheta \rightarrow Y$
\State Use adaptive function based on local image variance $\varrho(x) = \exp(-\psi \cdot {Var}(x))$

\For{each cell $C_i$ in $\vartheta$}
    \State Compute the adaptive weighting from $w_i = \varrho(x_i)$
    \State Match $w_i$ from region $C_i$ to the corresponding weight at $y_i$
\EndFor

\State Apply the cost function for contextual sensitivity: $
c_C(x, y; C) = c(x, y) \times (1 + \upsilon \cdot e^{-\tau \cdot \| C(x) - C(r_0) \|})$

\State Calculate the total cost of 

$f(x)$: $f(x) = \sum_{i \in \mathcal{I}} \int_{C_i} c_C(x, f(x)) d\varepsilon(x)$

\State Use $g_\theta^{(t)}(x)$ to refine intermediate reconstructions at each timestep $t$
\For{$t=T$ to $1$}
    \State Compute the reconstructed image $\hat{x}_t = g_\theta^{(t)}(x_t)$
    \State Update $f$ by minimizing the total cost using $w_i$ and $\hat{x}_t$
\EndFor

\State \Return Transport map minimizing CAD (use as input for Algorithm 2)
\end{algorithmic}
\end{algorithm}
%
\noindent in which \( \mu_\theta \) and \( \sigma_\theta \) denote the learned parameters of the model.
In pixel-wise comparison, current diffusion models typically use the Euclidean distance to measure the discrepancy between the generated \( \hat{x}_t \) and the reference \( x_t \) images as follows,
\begin{equation}
\begin{array}{lr}
\begin{aligned}
\mathcal{L}_{diffusion} = \mathbb{E}_{q(x_{1:T} | x_0)} \left[ \sum_{t=1}^{T} \| x_t - \hat{x}_t \|_2^2 \right].
\end{aligned}
\end{array}
\label{4-4}
\end{equation}
While Euclidean distance is simple and computationally efficient, it has limitations for image completion: it treats all pixel differences equally, regardless of their spatial arrangement or context, leading to minimized local pixel values but compromised overall perceptual quality and coherence. Additionally, Euclidean distance fails to capture complex structural and semantic relationships within an image, overlooking higher-level features like edges, textures, and object boundaries (see Fig. \ref{fig2}). It also does not consider the broader context, leading to inconsistencies where generated pixels appear out of place with their surroundings.

Another approach to addressing this limitation is the Wasserstein Discrepancy (WD) \cite{panaretos2019statistical}. At the core of this method is the gradient of the Brenier potential, which defines the optimal transport map for transforming the degraded image distribution into its restored form with minimal cost. This alignment helps capture the underlying geometric structure, ensuring that the generated content is consistent with the reference image. WD is defined as:
\begin{equation}
\begin{array}{lr}
\begin{aligned}
W(\mathbb{P}, \mathbb{Q}) = \inf_{\gamma \in \Pi(\mathbb{P}, \mathbb{Q})} \mathbb{E}_{(x, y) \sim \gamma} [\| x - y \|],
\end{aligned}
\end{array}
\label{4-5}
\end{equation}
\noindent in which \( \mathbb{P} \) and \( \mathbb{Q} \) denote probability distributions, and \( \Pi(\mathbb{P}, \mathbb{Q}) \) represents the set of all distributions that have \( \mathbb{P} \) and \( \mathbb{Q} \) as their marginal distributions. As shown in Fig. \ref{fig2}, while the WD outperforms the Euclidean distance by considering spatial relationships and the distribution of pixel values, it still does not minimize the discrepancy between the generated and reference images effectively. This indicates that although WD captures geometric structure, it does not fully address factors such as local coherence or semantic alignment. 

\begin{figure}[t]
\centering
\includegraphics[width=0.96\columnwidth]{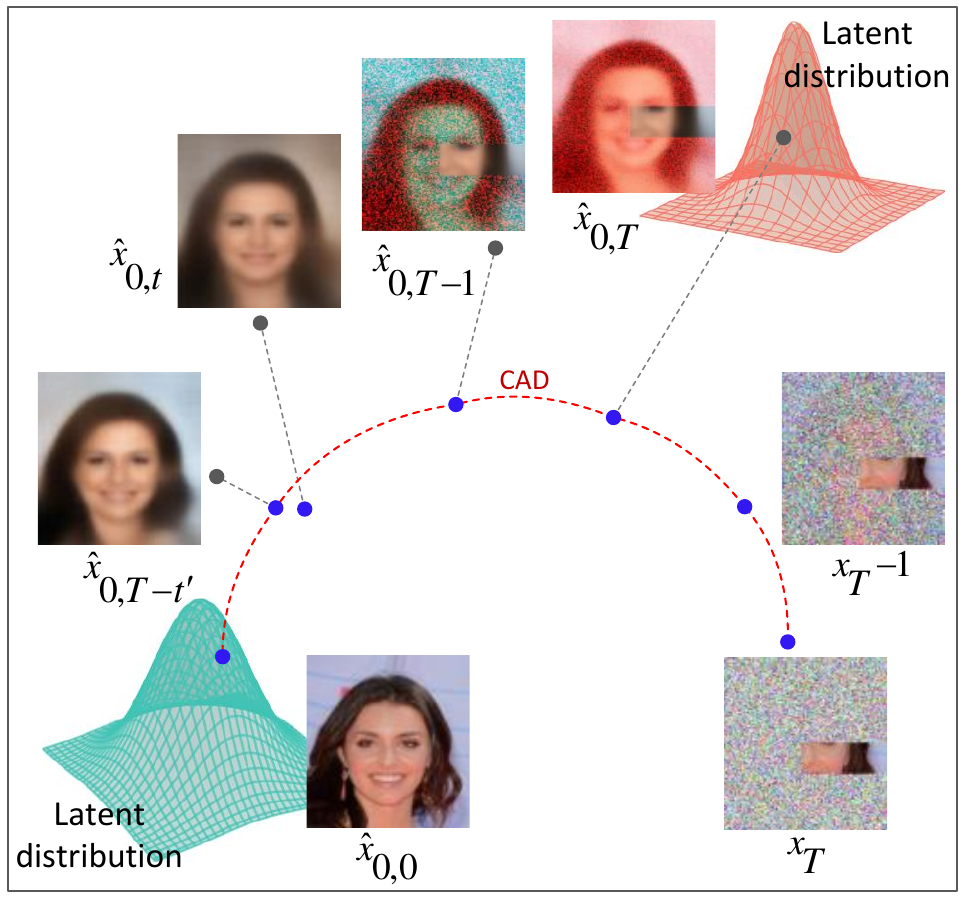} 
\vspace{-6pt}
\caption{\small{The ConFill framework. The red curve represents the CAD inverse diffusion steps, which iteratively balance the distribution of image patches across the latent space. The green and red hills represent latent distributions closer to the final completed image and noisy initial steps, respectively.}}
\label{arch}
\end{figure} 

\vspace{-3pt}
\subsection{Foundational Strategy}
\label{s:4.2}
To address the above limitations, we introduce CAD, which integrates contextual information into the discrepancy measure as
\begin{equation}
\begin{array}{lr}
\begin{aligned}
W_C(x, y; C) = \|\nabla \phi(x) - y\|^2\cdot f(C(x), C(y)),
\end{aligned}
\end{array}
\label{4-6}
\end{equation}
in which \( C(\cdot) \) denotes contextual features extracted from the image, such as texture or color gradients, 
\( \phi(x) \) is the Brenier potential that defines an optimal transport map to ensure monotonic and cost-efficient displacement of probability mass between latent distributions. The discrepancy function adapts dynamically based on local and global features, aligning the generative process to a transport-based regularization, and \( f(\cdot, \cdot) \) is a scaling function based on the differences between these features
\begin{equation}
\begin{array}{lr}
\begin{aligned}
f(C(x), C(y)) = 1 + \upsilon \cdot e^{-\tau \cdot \|C(x) - C(y)\|},
\end{aligned}
\end{array}
\label{4-7}
\end{equation}
\noindent where \( \upsilon \) and \( \tau \) mainly control the sensitivity of the scaling function to contextual differences. This process ensures that the discrepancy metric dynamically adjusts to both local and global image features, providing a more accurate measure of similarity.
Therefore, the CAD loss can be written as
\begin{equation}
\begin{array}{lr}
\begin{aligned}
\mathcal{L}_{CAD} = \mathbb{E}_{q(x_{1:T} | x_0)} \left[ \sum_{t=1}^{T} W_C(\hat{x}_t, x_0; C) \right].
\end{aligned}
\end{array}
\label{4-8}
\end{equation}
In our approach, we refine \( x_t \) by minimizing the CAD loss
\begin{equation}
\begin{array}{lr}
\begin{aligned}
 \resizebox{.97\hsize}{!}{$\mathcal{L}_{CAD} = \| x_t - \mu_\theta(x_{t+1}, t+1) \|_2^2 + \gamma_t \cdot W_C(s(g_\theta(x_t)), r_0; C)$},
\end{aligned}
\end{array}
\label{4-9}
\end{equation}
\noindent where \( \gamma_t \) is a weighting parameter, and \( g_\theta \) is the denoising function. Then apply gradient descent to update \( x_t \) at each timestep ($x_{t-1} = x_t - \lambda \cdot \nabla_{x_t} \mathcal{L}_{CAD}$), where \( \lambda \) is the learning rate.
Considering the diffusion process for all \( t \), the denoising follows a deterministic process aimed at minimizing CAD between \(\hat{x}_T\) and the original data distribution, aligning the model with the target distribution. Therefore, CAD directly influences the optimization objective in Eq. (\ref{4-9}) and estimation of \(\hat{x}_0\).
Thus, the completion constraint $\mathcal{O}:s(\hat{x}_0)$, can be applied to \(\hat{x}_T\). Therefore, addressing the image completion challenge depends on determining an optimal \(\hat{x}_T\) guided by the posterior distribution
\begin{equation}
\begin{array}{lr}
    p_{\theta}\left(\hat{x}_T \mid \mathcal{O}\right) = p_{\theta}\left(\hat{x}_T\right) \cdot \delta_{{CAD}}\left(s(\hat{x}_0) = r_0\right),
    \label{4-10}
\end{array}
\end{equation}
\noindent where \(\delta_{{CAD}}\) is the delta function in the context of CAD, emphasizing the optimal alignment of \(\hat{x}_0\) and \(r_0\).
As established in Eq. (\ref{4-2}) and Eq. (\ref{4-3}), \( p_{\theta}\left(\hat{x}_T\right) \) is a Gaussian distribution. Notably, \( p_{\theta}\left(s(\hat{x}_0) = r_0 \mid \hat{x}_T\right) \) represents the conditional likelihood of \( s(\hat{x}_0) \) at \( r_0 \), given \( \hat{x}_T \). With \( \hat{x}_T \) fixed and \( \hat{x}_0 \) as a deterministic function of \( \hat{x}_T \), this measure is represented by a delta function \( \delta_{{CAD}}(\cdot) \), which reflects the alignment between \( s(\hat{x}_0) \) and \( r_0 \). This function indicates perfect alignment, where the CAD is minimized, ensuring that the transformation between image distributions more effectively minimizes discrepancies. Consequently, the logarithm of the posterior distribution in Eq. (\ref{4-10}) can be directly estimated as follows
\begin{equation}
\begin{array}{lr}
    \resizebox{.97\hsize}{!}{$\log p_{\theta}\left(\hat{x}_T \mid \mathcal{O}\right) \approx -\frac{1}{2}\left\|\hat{x}_T\right\|_{2}^{2}-\frac{1}{2 \gamma_{T}^{2}}{CAD}\left(r_0, s(\mathcal{F}_{\sigma}(\hat{x}_T))\right)^{2} + \Gamma$},
    \label{4-11}
\end{array}
\end{equation}
\noindent where \(\mathcal{F}_{\sigma}(\hat{x}_T) = \hat{x}_0\) represents that \(\hat{x}_0\) is functionally dependent on \(\hat{x}_T\). The \(\Gamma\) denotes the normalising constant, while \(\gamma_{T}\) represents the dispersion of the distribution, shaping the precision of the CAD-based approximation. As \(\gamma_{T}\) tends towards zero, the approximation in Eq. (\ref{4-11}) becomes increasingly accurate. Indeed, optimally selecting \(\gamma_{T}\) can significantly enhance performance.
Eq. (\ref{4-11}) provides a basis for determining \(\hat{x}_T\) using the optimization method. The first part of this equation serves as a guiding framework, while the second part imposes a constraint to ensure accurate completion. To estimate \(\hat{x}_T\), we use wasserstein gradient flows \cite{mokrov2021large} to generate random samples from complex distributions. This optimization effectively addresses the inconsistency issues observed in current methods \cite{fan2021variational}. 
\vspace{2pt}

\noindent {\bf {Contextual Feature Extractor (\(C(\cdot)\))}}: To compute the context-adaptive discrepancy \(c_C(x, y; C)\), the \(C(\cdot)\) combines handcrafted and learned features. I) Handcrafted features are used to capture low-level structural and color information. These include edge-based features, texture-based features, and color distributions. Together, these features provide a robust representation of local structural patterns and variations in color across the image. II) Learned features are used to extract high-level semantic features, including object boundaries, relationships, and contextual patterns.
Our workflow is illustrated in Fig. \ref{arch}.
In our model Algorithm 1 computes the transport map \( f \) by minimizing the CAD loss. This ensures alignment between the generated regions and the known parts of the image using adaptive weightings and cost functions. \( f \) serves as an intermediate representation and acts as input to Algorithm 2. This Algorithm takes \( f \) along with \( g_\theta \) to iteratively refine the image reconstruction.


\begin{algorithm}[t]\small
\caption{ConFill}
\label{alg2}
\begin{algorithmic}[1]
\Require $f$ (from Algorithm 1), $r_0$, $g_\theta$, $\acute{\mathfrak{t}}$, $\mathcal{J}$, $G$, and $\lambda_t$ (at $t$)

\State \textbf{Initialize:} $\hat{x}_T \sim \mathcal{N}(0, \text{I})$
\State Set $t = T$ and $j = \mathcal{J}$

\While{$t > 0$}
    \State \textbf{Step 1: CAD Refinement}
    \State \quad Compute transport potential: $\nabla \phi_t(\hat{x}_t)$ (Eq. (\ref{4-6}))
    \State \quad Update $\hat{x}_t$ by minimizing: $
    \text{CAD}(\hat{x}_t, r_0) + \lambda_t ||\nabla \phi_t(\hat{x}_t) - \hat{x}_t||$ (Eq. (\ref{4-9}))
    \State \quad Apply gradient penalty to enforce Lipschitz constraints.
    
    \State \textbf{Step 2: Reverse Diffusion Update}
    \State \quad Compute $\hat{x}_{t-1}$ via CAD-driven transition: $
    \hat{x}_{t-1} = g_\theta(t)(\hat{x}_t) - \lambda \nabla \phi_t(\hat{x}_t)$
    \State \quad Reduce timestep: $t \leftarrow t - 1$
    
    \State \textbf{Step 3: Temporal Regression}
    \If{$t \leq T - \acute{\mathfrak{t}}$ \textbf{and} $t \bmod \acute{\mathfrak{t}} = 0$}
        \If{$j > 0$}
            \State Predict $\hat{x}_{t+\acute{\mathfrak{t}}}$ via transition distribution $q(\hat{x}_{t+\acute{\mathfrak{t}}}|\hat{x}_t)$ (Eq. (\ref{eq:17}))
            \State Adjust timestep: $t \leftarrow t + \acute{\mathfrak{t}} - 1$
            \State Decrement jump count: $j \leftarrow j - 1$
        \Else
            \State Reset jump count: $j \leftarrow \mathcal{J}$
        \EndIf
    \EndIf
\EndWhile
\State \textbf{Output:} Reconstructed image $\hat{x}_0$
\end{algorithmic}
\end{algorithm}

One of the challenges with our approach is its computational demand. Specifically, generating the final output \( \mathcal{F}_{\sigma}\left(\hat{x}_{T}\right) \) and calculating its gradient require executing both forward and backward passes through a denoising sequence. To address this, it is essential to improve the computational efficiency of the algorithm while preserving the efficacy of our strategy.
Referring to the discussion in Sec. \ref{s:4.2}, a practical modification involves substituting the complex evaluation of \( {\mathcal F}_{\sigma}\left(\hat{{x}}_{T}\right) \) in Eq. (\ref{4-11}) with the gradual generation \( {g}_{\theta}^{(T)}\left(\hat{{x}}_{T}\right) \). This adjustment expedites the approximation of the final output and streamlines the computation process.
Therefore, we establish an estimated conditional distribution model for \(s(\hat{x}_0)\) conditional on \(\hat{x}_T\), represented as \(p_{\theta}^{\prime}(s(\hat{x}_0) | \hat{x}_T)\). This estimation is produced by a singular step generation, \(s(g_{\theta}^{(T)}(\hat{x}_T))\),  and incorporates a small Gaussian perturbation to model the variability.
\begin{equation}
\begin{array}{lr}
    p_{\theta}^{\prime}(\hat{x}_T | s(\hat{x}_0))  \sim \mathcal{N}(s(\hat{x}_0); s(g_{\theta}^{(T)}(\hat{x}_T)), \gamma_{T}^{\prime 2} \text{I}),
    \label{eq:12}
\end{array}
\end{equation}
\noindent in which $\gamma_{T}^{\prime}$ denotes the parameter of standard deviation. Upon integrating this estimated distribution, we obtain an accurate approximation of the posterior distribution
\begin{equation}
\begin{array}{lr}
    \resizebox{.97\hsize}{!}{$\log p_{\theta}^{\prime}\left(\hat{x}_{T} \mid \mathcal{O}\right) \approx -\frac{1}{2}\left\|\hat{x}_{T}\right\|_{2}^{2}-\frac{1}{2 \gamma_{T}^{\prime 2}}{CAD}\left(r_{0}, s\left(g_{\theta}^{(T)}\left(\hat{x}_{T}\right)\right)\right)^{2}
    + \Gamma^{\prime}$},
    \label{eq:13}
\end{array}
\end{equation}
\noindent in which $\Gamma^{\prime}$ acts as a normalisation factor.
To effectively reduce the approximation gap, between \(p_{\theta}(s(\hat{x}_{0}) | \hat{x}_{T})\) and \(p_{\theta}^{\prime}(s(\hat{x}_{0}) | \hat{x}_{T})\), we propose adjusting the variance parameter \( \gamma_{T}^{\prime 2} \) with respect to CAD. This ensures the estimated distribution more accurately reflects the true distribution, thereby enhancing the approximation.
\begin{equation}
\begin{array}{lr}
    \gamma_{T}^{\prime 2} = \frac{1}{\mathscr{N}} \sum_{i=1}^{\mathscr{N}} {CAD}\left(s\left(g_{\theta}^{(T)}(\hat{x}_{T}^{(i)})\right), s(\hat{x}_{0}^{(i)})\right),
    \label{eq:14}
\end{array}
\end{equation}
\noindent we adjust \(\gamma_{T}^{\prime 2}\) based on the average CAD across samples, where \(\mathscr{N}\) represents the $r_{0}$'s dimension. Here, \({CAD}(\cdot, \cdot)\) measures the shortest distance between the transformed versions of \(\hat{x}_{T}\) and \(\hat{x}_{0}\).
Maximizing Eq. (\ref{eq:13}) with respect to \( \hat{x}_T \) aims to balance the image completion constraint with prior regularization. Indeed, unlike Eq. (\ref{4-11}), which minimizes \( \gamma_T \), Eq. (\ref{eq:14}) adjusts \( \gamma_T' \) to be sufficiently large. This adjustment minimises the approximation error, thereby reducing the emphasis on the image completion constraint in Eq. (\ref{eq:13}). Consequently, this ensures a balanced approach to minimizing errors while maintaining image integrity.
\vspace{-3pt}
\subsection{Dynamic Sampling}
To enhance the effect of CAD in regions with varying complexity, we introduce a novel dynamic sampling technique. This strategy dynamically modulates the sampling density by using an adaptable function, which evaluates the textural and structural complexity inherent in each image region. Textural complexity is quantified as the local variance of pixel intensities within a region as $\mathrm{Var}(x) = \frac{1}{N} \sum_{i=1}^N (x_i - \mu)^2,$ where $\mu = \frac{1}{N} \sum_{i=1}^N x_i$, and  \(x_i\) represents the pixel intensities.
Moreover, the structural complexity is measured using edge density (ED), defined as the proportion of edge pixels in the region. The two measures are combined into the adaptability function \( \varrho(x) \), which assigns weights based on the relative contributions of textural and structural complexity:
\begin{equation}
\begin{array}{lr}
    \varrho(x) = \hat{\alpha} \cdot \exp\left(-\psi \cdot \frac{\mathrm{Var}(x) \cdot N}{N-1}\right) + \hat{\beta} \cdot \text{ED}(x),
    \label{eq:as1}
\end{array}
\end{equation}
where $\hat{\alpha}$ and $\hat{\beta}$ control the relative importance of textural and structural complexity, \( {Var}(x) \) represents the local variance within a specific region of the image, and $\psi$ is a scaling parameter that adjusts sensitivity to textural variations.
The dynamic sampling process enhances CAD computation by integrating the adaptability function as a weighting factor within the discrepancy measure. Consequently, CAD can be computed as
\begin{equation}
\begin{array}{lr}
    {\mathrm{CAD}}(x, y) = \sum_{i=1}^{N} \varrho(x_i) \cdot W(p_{x_i}, p_{y_i}),
    \label{eq:as2}
\end{array}
\end{equation}
where \( W(p_{x_i}, p_{y_i}) \) is the discrepancy between the distributions \( p_{x_i} \) and \( p_{y_i} \) at points \( x_i \) and \( y_i \).
During the image completion process, the sampling rate for each pixel or region is adjusted based on \( \varrho(x) \), effectively increasing the density of sample points in regions with higher complexity $\mathcal S(x) = \lceil \mathfrak{m} \cdot \varrho(x) \rceil$,
where $\mathcal S(x)$ represents the number of samples, and $\mathfrak{m}$ is the maximum samples per region.
By dynamically adjusting the sampling based on local image characteristics, CAD enables more precise control over the restoration process. 
The CAD framework balances local and global features by capturing edges and textures through the handcrafted component of \( C(\cdot) \) and ensuring global semantic consistency using high-level features from a pre-trained network. This dual strategy dynamically adapts the discrepancy metric, aligning fine-grained details with broader structural coherence for superior completion quality.

\vspace{-5pt}
\subsection{Gradual Approximation}
To further refine our stochastic diffusion process, we gradually refine the approximation error at each stage of the process. This ensures that the approximated generation closely aligns with the reference image \( r_0 \).
Our diffusion model draws samples \(\hat{x}_{0:T}\) from an estimated posterior distribution \(p_{\theta}^{\prime}(\hat{x}_{0:T} | \mathcal{O})\), in order to reduce the gap between the estimated distribution and the true distribution of \(\hat{x}_{0:T}\) given \(\mathcal{O}\), which can be expressed as
\begin{equation}
\begin{array}{lr}
    p_{\theta}^{\prime}\left(\hat{{x}}_{0: T} \mid \mathcal{O}\right)=p_{\theta}^{\prime}\left(\hat{{x}}_{T} \mid \mathcal{O}\right) . \prod_{t=1}^{T} p_{\theta}^{\prime}\left(\hat{{x}}_{t-1} \mid \hat{{x}}_{t}, \mathcal{O}\right),
    \label{eq:15}
\end{array}
\end{equation}

\noindent in which \(p_{\theta}^{\prime}\left(\hat{{x}}_{T} \mid \mathcal{O}_{T}\right)\) is defined based on minimising the CAD, as indicated in Eq. (\ref{eq:13}). To compute \(p_{\theta}^{\prime}\left(\hat{{x}}_{t-1} \mid \hat{{x}}_{t}, \mathcal{O}\right)\), we apply a series of Gaussian approximation distributions to minimize the distance to the true conditional distributions at each step \(t\), similar to the approach described in Eq. (\ref{eq:12}).
\begin{equation}
\begin{array}{lr}
    p_{\theta}^{\prime}\left(\hat{{x}}_{t} \rightarrow {s}(\hat{{x}}_{0})\right) \sim \mathcal{N}\left({\mathrm{CAD}}({g}_{\theta}^{(t)}\left(\hat{{x}}_{t}\right) \rightarrow {s}(\hat{{x}}_{0})), \text{I} \cdot \gamma_{t}^{\prime 2}\right).
    \label{eq:16}
\end{array}
\end{equation}
For minimising the gradual approximation error, we define \(\gamma_{t}^{2}\) as per Eq. (\ref{eq:14}), with \(t\) replacing \(T\). This adjustment allows us to more accurately compute \(p_{\theta}^{\prime}(\hat{x}_{t-1} | \hat{x}_{t}, \mathcal{O})\) as
\begin{equation}
\begin{array}{lr}
\begin{aligned}
    & \log p_{\theta}^{\prime}\left(\hat{x}_{t-1} \mid \hat{x}_{t}, \mathcal{O}\right)
\approx -\frac{1}{2 \sigma_{t}^{2}}\left\|\hat{x}_{t-1}-\hat{\kappa}_{t}\right\|_{2}^{2} \\
& -\frac{1}{2 \gamma_{t-1}^{\prime 2}}{\mathrm{CAD}}\left(r_{0}-s\left(g_{\theta}^{(t-1)}\left(\hat{x}_{t-1}\right)\right)\right)^{2} +\Gamma^{\prime},
\end{aligned}
    \label{eq:17}
\end{array}
\end{equation}
\noindent where the derivation of the second term can be written as
\begin{equation}
\begin{array}{lr}
\begin{aligned}
   & \hat{\kappa}_{t} = g_{\theta}^{(t)}(\hat{x}_{t}) \cdot \sqrt{\alpha_{t-1}} \\
   & + \left( \frac{\hat{x}_{t} - g_{\theta}^{(t)}(\hat{x}_{t}) \cdot \sqrt{\alpha_{t}}}{\sqrt{1-\alpha_{t}}} \right) \cdot \sqrt{1-\alpha_{t-1}-\sigma_{t}^{2}},
\end{aligned}
    \label{eq:18}
\end{array}
\end{equation}

\noindent where $\alpha_t=1-\beta_t$ denotes the noise level at each timestep and the standard deviation of the noise distribution is $\sigma$. For generating the final completion result, we use a step-by-step optimization method aimed at selecting sequences of \(\hat{x}_{0:T}\) that increase the probability \(p_{\theta}^{\prime}(\hat{x}_{0:T} | \mathcal{O})\) mentioned in Eq. (\ref{eq:15}). This process starts by choosing an initial \(\hat{x}_{T}\) based on Eq. (\ref{eq:13}). Using \(\hat{x}_{T}\), the subsequent sample \(\hat{x}_{t-1}\) is obtained by optimizing the objective function outlined in Eq. (\ref{eq:17}). These iterative steps are designed to satisfy the completion criteria within the denoising diffusion approach.
Fig. \ref{arch} shows that the CAD-based approximation error decreases as \(t\) reduces, indicating that the algorithm progressively improves fidelity to the completion constraint with each step. Specifically, when \(t\) reaches 1, by assigning \(\sigma_{1}=0\) and reducing \(\gamma_{1}\) towards zero, we can minimise the approximation error to an extent where \(g_{\theta}^{(1)}(\hat{x}_{1})\) approximates \(\hat{x}_{0}\) with high accuracy.
This method ensures high quality generation while efficiently reducing transport costs with each iteration.

While our algorithm reduces approximation error during the final denoising phase, early inaccuracies still affect image quality by influencing the prior distribution for later steps. To address this, during the initial denoising phase, in which the significance of approximation errors is higher, we adopt a multi-step approximation approach. This involves approximating \(\hat{x}_{0}\) by traversing through multiple deterministic denoising steps at selected time intervals.
As outlined in \cite{lugmayr2022repaint, wang2022zero, zhang2023towards}, we can integrate the time travel method to bolster the self-consistency of intermediate samples.
This method involves returning previous denoising steps by adding noise to the intermediate images. 
At the denoising time step $T - \acute{\mathfrak{t}}$, we consider a predefined range of time steps denoted as $\rho$. Indeed, rather than proceeding to $T - \acute{\mathfrak{t}} - 1$, we backtrack to time $T - 1$ and sample a new $\hat{x}_{T-1}$ according to $q(\hat{x}_{T-1} | \hat{x}_{t-\acute{\mathfrak{t}}})$, and repeat the denoising steps from that point. This iterative process repeats for $\mathcal{J}$ rounds, alternating between rewinding and denoising. The sequence progresses through steps $T-1$ to $T-\acute{\mathfrak{t}}$ and then $T-\acute{\mathfrak{t}}-1$ to $T-2\acute{\mathfrak{t}}$ repeating iteratively until reaching time zero. The details of our model's procedure are illustrated in Algorithm \ref{alg2}.

\begin{figure}
\centering
\includegraphics[width=0.99\columnwidth]{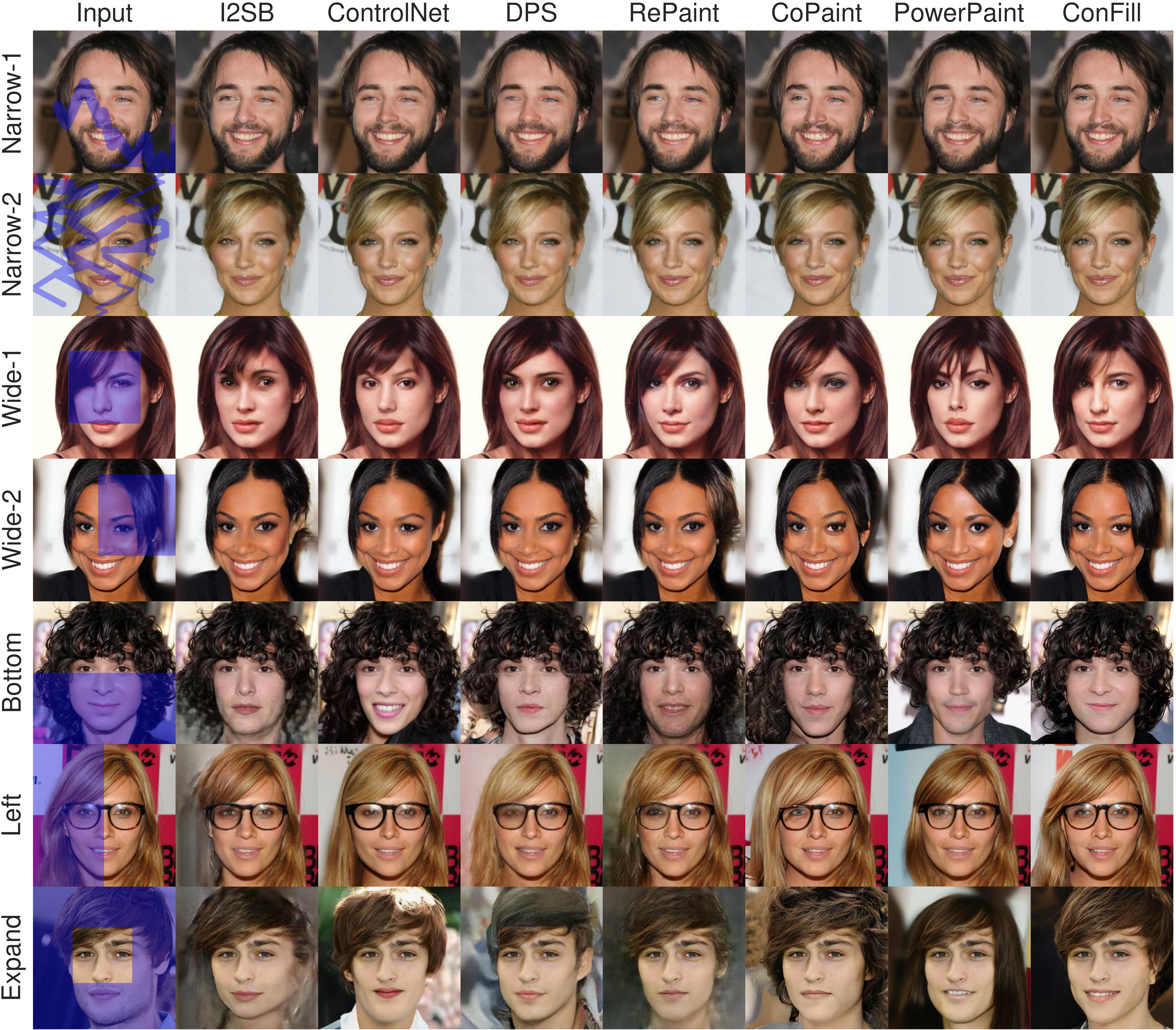} 
\vspace{-4pt}
\caption{\small{Qualitative comparison on the CelebA-HQ. The facial images generated by ConFill not only present more distinctive facial features but also show a higher degree of similarity and coherence with the original images compared to those produced by other baseline models.}}
\label{celeba}
\end{figure} 

\vspace{-3pt}
\section{Experiments}
\subsection{Setup Details}
{{\bf{Datasets and models:}} Our methodology is evaluated on three widely recognised image datasets: CelebA-HQ \cite{liu2015deep}, Places2 \cite{zhou2017places}, and ImageNet-1K \cite{russakovsky2015imagenet}. Given that diffusion models typically require square-shaped images, we pre-process the datasets by cropping all images to a uniform size of $256\times256$ pixels to ensure compatibility with pre-trained diffusion models. We use the pre-trained diffusion model from \cite{lugmayr2022repaint} and \cite{dhariwal2021diffusion} for the CelebA, Places2 and ImageNet. Hyperparameter selection is conducted using the first five images from the validation set. Evaluation is performed on the images from the test sets, similar to the methodology outlined by \cite{lugmayr2022repaint}. 
For the contextual feature extractor \(C(\cdot)\) we use Sobel and Gabor filters for edge and texture extraction. Additionally, a ResNet18 is used to extract high-level semantic features. The network is pre-trained on ImageNet and fine-tuned on the target dataset.
\vspace{2pt}

\noindent {\bf{Masks:}}  We use masks from \cite{suvorov2022resolution} and \cite{lugmayr2022repaint}. These masks are created by uniformly sampling from polygonal chains that have been expanded by a randomly determined width, combined with rectangles of diverse aspect ratios and orientations.
\vspace{2pt}

\noindent {\bf{Metrics:}} The quality of the completion results is evaluated through a combination of objective and subjective metrics. We use the Fréchet Inception Distance (FID), Learned Perceptual Image Patch Similarity (LPIPS), and Structural Similarity Index Measure (SSIM). For each image, two completed images are produced, and we report the mean scores. 
Our evaluation focuses on the degree to which the completed images achieve a natural appearance free of artefacts, closely mirror the original reference image in the areas that were not masked, and maintain a cohesive visual flow throughout the entire image. This set of criteria is designed to assess the quality of the completion results, focusing on realism, fidelity to the original scenes, and internal consistency. 
\begin{figure}
\centering
\includegraphics[width=0.99\columnwidth]{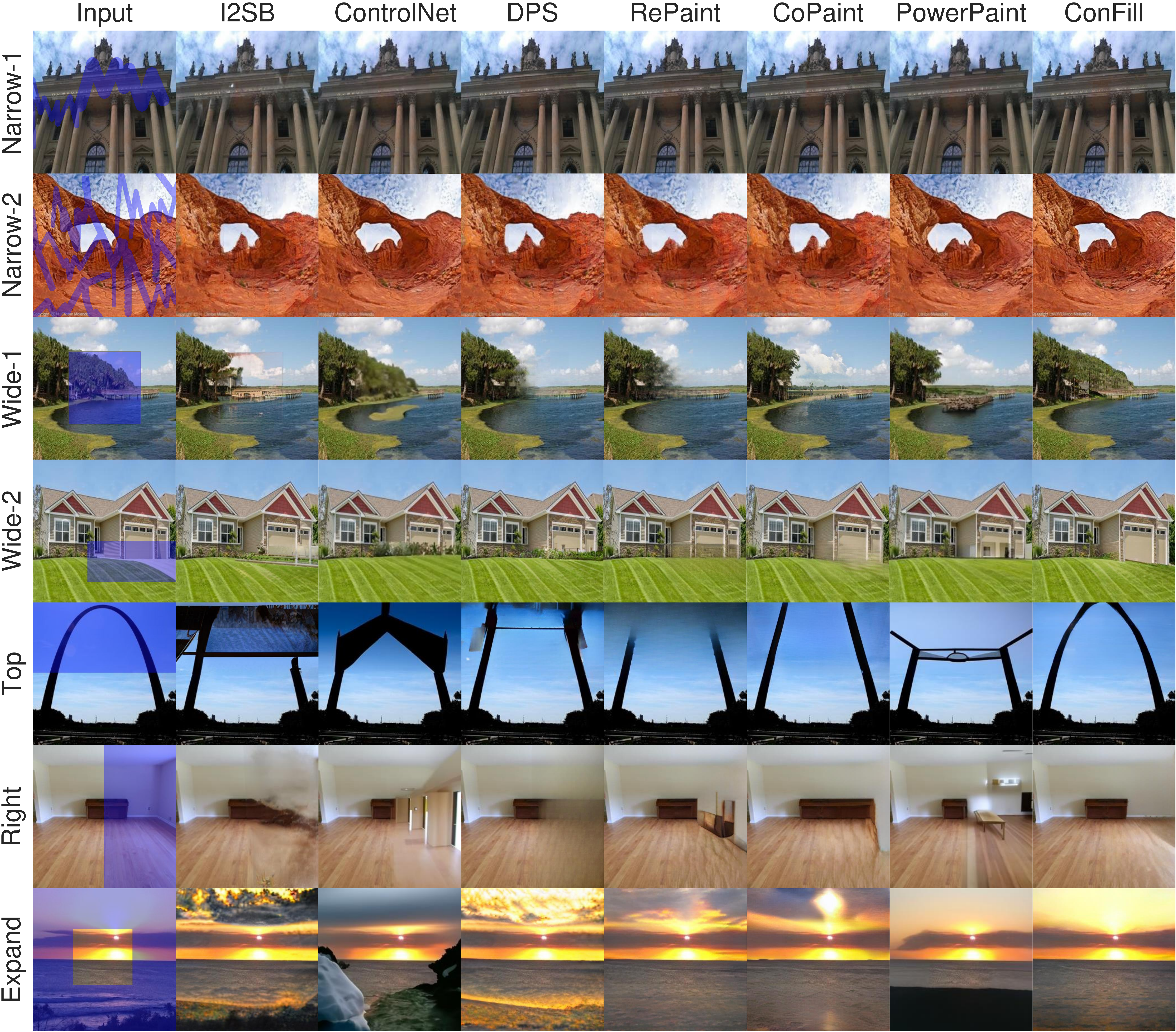} 
\vspace{-4pt}
\caption{\small{Qualitative comparison on the Places2. ConFill outperforms the other models, providing accurate and seamless restoration. The inpainted regions blend naturally with the surrounding landscape, maintaining consistent texture and structure across the images.}}
\label{place}
\end{figure} 

\vspace{2pt}
\noindent {\bf{Baselines comparisons and implementation details: }}
Our analysis focuses on diffusion approaches, which have demonstrated superior performance in comparison to non-diffusion model approaches \cite{lugmayr2022repaint}. To this end, we evaluate against established baselines, including DDRM \cite{kawar2022denoising}, RePaint \cite{lugmayr2022repaint}, CoPaint \cite{zhang2023towards}, DPS \cite{chung2022diffusion}, I$^{2}$SB\cite{liu20232}, ControlNe \cite{zhang2023adding}, and PowerPaint \cite{zhuang2025task} to highlight our comparative study. We follow the original implementations for selecting the NFE and hyper-parameters for each method.

Our model has been developed using PyTorch, and we use four RTX 3080-TI GPUs for training and conducting all experiments. In the case of RePaint, we applied the publicly available code, following the configurations outlined in the original paper. We adapted the implementation of all other methods from RePaint, ensuring that the hyperparameters remained consistent with those detailed in their papers. For ConFill, specifically, we set the number of gradient descent steps, $G=2$. We apply CAD to balance the initial token distribution before generating the final completed image. Additionally, we alternatively applied it to the intermediate steps to refine token assignments.
Here, \(\hat{x}_{t}\) is iteratively optimised to maximise the posterior probability at timesteps. This process is equivalent to minimising the loss for \(\hat{x}_{t}\).
\begin{figure}
\centering
\includegraphics[width=0.99\columnwidth]{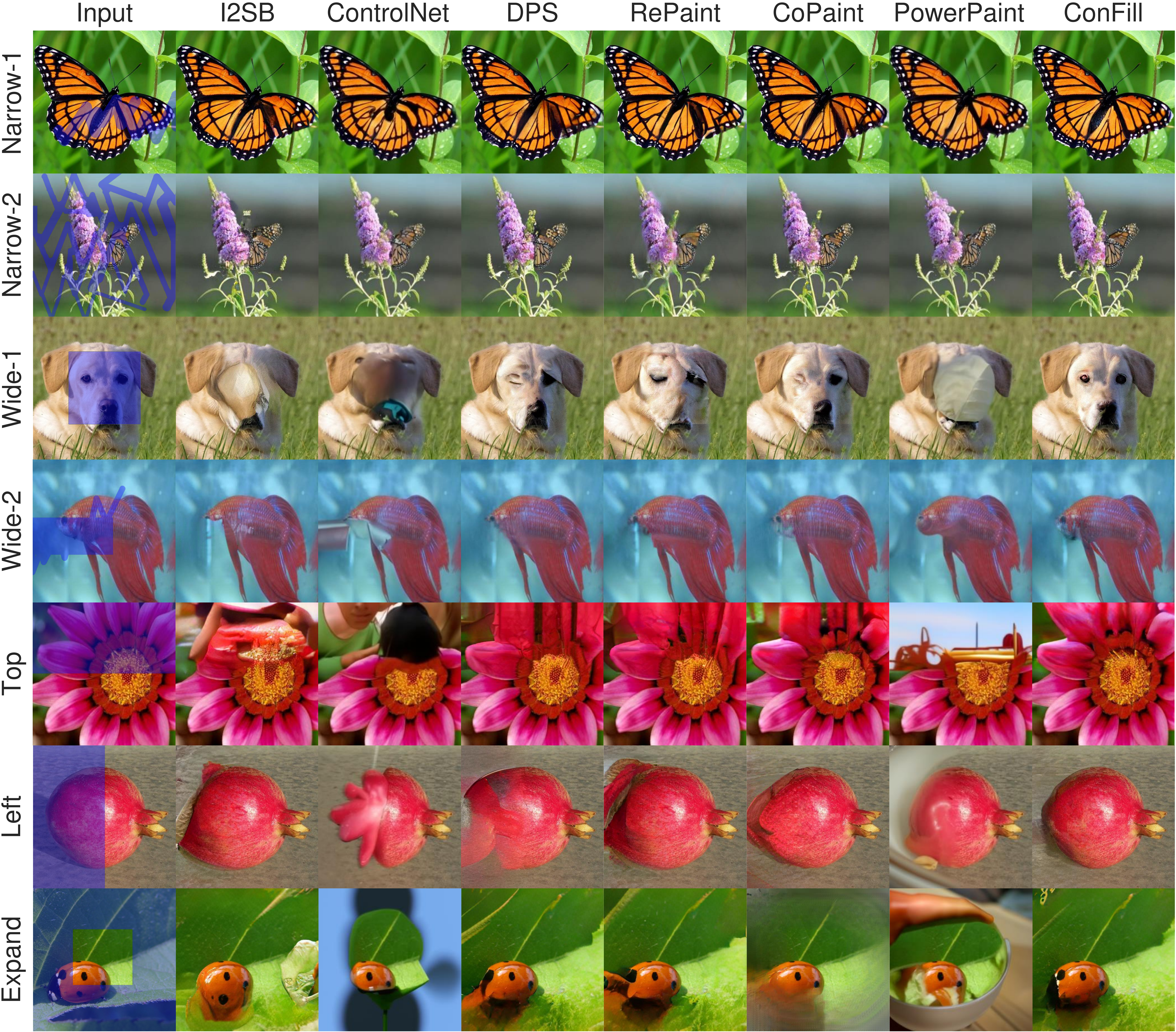}
\vspace{-4pt}
\caption{\small{Qualitative comparison on the ImageNet-1k. ConFill outperforms other methods in terms of preserving structural and details.}}
\label{imagenet}
\end{figure} 
\begin{table*}
\centering
\caption{\small{Quantitative comparisons on three (CelebA, Places2, and ImageNet-1k) datasets with various mask types. Results are based on FID ($\downarrow$), LPIPS ($\downarrow$) and SSIM ($\uparrow$). For each mask type and metric, {\bf{bold}} indicates the best result and \underline{under} the second best.}}
\vspace{-4pt}
\label{tab.1}
\resizebox{18.2cm}{!}{
\begin{tabular}{@{}c| c c c|c c c| c c c| c c c| c c c|c c c|c c c@{}}
\hline 
\multirow{3}{*}{Mask} & \multicolumn{3}{c|}{ConFill (ours)} & \multicolumn{3}{c|}{PowerPaint}&\multicolumn{3}{c|}{I$^{2}$SB} & \multicolumn{3}{c|}{DPS} & \multicolumn{3}{c|}{RePaint} & \multicolumn{3}{c|}{CoPaint} & \multicolumn{3}{c}{ControlNet}  \\ \cline{2-22}
&FID & LPIPS & SSIM   & FID & LPIPS & SSIM & FID & LPIPS & SSIM & FID & LPIPS & SSIM & FID & LPIPS & SSIM & FID & LPIPS & SSIM & FID & LPIPS & SSIM  \\ 
& \multicolumn{21}{c}{\cellcolor{gray}{\textcolor{white}{\it{CelebA-HQ}}}}   \\ \cline{1-22}


Narrow & \bf{7.08} & \bf{0.034} & \bf{0.896} &8.65 &0.058 &0.859 & 8.03 & 0.051 & 0.864&8.72 &0.063 &0.854 &\underline{7.32} &0.045 &0.881 &7.39 &\underline{0.041} & \underline{0.887}&8.77 &0.049 &0.851  \\
 
Wide1 & \bf{8.05} & \bf{0.064} & \bf{0.862} &9.18 &0.097 &0.810 &8.79 & 0.087 & 0.835&9.26 &0.085 &0.821 & 8.54 &0.077 &\underline{0.857} &\underline{8.48} &\underline{0.070} &0.851 &8.93 &0.091 &0.818  \\

Wide2 & \bf{8.13} & \underline{0.076} & \bf{0.854} &9.36 &0.102 &0.804 & 8.85 & 0.095 & 0.830&9.31 &0.093 &0.824 & \underline{8.61} &\bf{0.073} & \bf{0.854} &8.70 &0.082 &\underline{0.842} &9.28 &0.097 &0.811  \\
 
Half ver. & \bf{12.94} & \bf{0.187} & \bf{0.735} &16.97 &0.236 &0.708 & 16.56 & 0.214 & 0.719&16.46 &0.198 &0.721 &\underline{16.38} &\underline{0.193} &\underline{0.725} & 16.49 & 0.203 & 0.720 &16.54 &0.205 &0.720 \\

Half hor. & \bf{12.67} & \underline{0.184} & \bf{0.742} &16.74 &0.229 & 0.716& 16.48 & 0.207 & 0.725&16.37 &0.192 &0.729 &\underline{16.34} &0.187 & 0.729 &\underline{16.34} &\bf{0.180} &\underline{0.735} &16.43 &0.210 & 0.727  \\

Expand & \bf{36.79} & \underline{0.440} & \bf{0.483} &54.22 &0.507 &0.447 & 53.51 & 0.498 & 0.443& {49.31} &\bf{0.435} & \underline{0.476} &{50.83} &0.476 &0.463 & \underline{48.75} & 0.472 &0.470 &52.61 &0.480 & 0.454 \\
\hline  
{Avg } & \bf{14.27} & \bf{0.164} & \bf{0.760} & 19.18 & 0.204 & 0.724 & 18.70 & 0.192 & 0.736 & 18.24 & 0.178 & 0.738 & 18.02 & 0.175 & 0.750 & \underline{17.69} & \underline{0.173} & \underline{0.751} &  18.76& 0.189 & 0.730  \\ 
\hline 
& \multicolumn{21}{c}{\cellcolor{gray}{\textcolor{white}{\it{Places2}}}}  \\ \cline{2-22}
 
{Narrow} & \bf{8.91} & \bf{0.063} & \bf{0.847} &10.54 &0.106 &0.805 & 10.56 & 0.094 & 0.817& 10.45 &0.108 &0.803 &{9.35} &0.072 &0.834 &\underline{9.26} &\underline{0.067} & \underline{0.841}&10.57 &0.108 &0.801  \\
 
 Wide1 & \bf{10.27} & \underline{0.110} & \bf{0.829} &12.20 &0.143 &0.773 & 12.03 & 0.123 & 0.804&12.02 &0.121 &0.796 & 11.73 &{0.116} &{0.820} &\underline{11.67} & \bf{0.106} &\underline{0.824} &12.13 &0.139 &0.780  \\
 
Wide2 & \bf{10.39} & \bf{0.120} & \bf{0.812} &12.49 &0.161 &0.770 & 12.13 & 0.138 & 0.797&12.23 &0.142 &0.789 & 12.05 & 0.128 & \underline{0.806} &\underline{11.86} &\underline{0.124} &0.802 &12.42 &0.154 &0.773  \\
 
Half ver. & \bf{14.02} & \bf{0.270} & \bf{0.663} &19.22 &0.290 &0.647 & 19.67 & 0.291 & 0.644&20.18 &0.306 &0.635 & {15.49} &{0.286} &{0.651} & \underline{15.41} & \underline{0.283}& \underline{0.658} &19.04 &0.289 &0.651 \\
 Half hor. & \bf{13.76} & \bf{0.258} & \bf{0.676} &19.34 &0.282 &0.641 & 19.86 & 0.284 & 0.651&20.09 &0.294 &0.643 &\underline{15.30} &0.269 & \underline{0.662} &{15.39} &\underline{0.264} &{0.660} &19.15 &0.278 & 0.647  \\
 
 {Expand} & \bf{48.65} & \bf{0.543} & \bf{0.417} &57.74 &0.602 &0.368 & 58.59 & 0.609 & 0.374&{57.83} &\underline{0.581} & \underline{0.388} &60.47 &0.618 &0.354 & \underline{57.47} & {0.591} &{0.378} &58.04 &0.611 & 0.359 \\
\hline  
{ Avg } & \bf{17.65} & \bf{0.227} & \bf{0.707} &21.93 & 0.263 & 0.667& 22.14 & 0.257 & 0.681 & 21.13 & 0.259 & 0.675 & 20.73 & 0.248 & 0.688 & \underline{20.17}& \underline{0.239} & \underline{0.694} & 21.89& 0.262 & 0.669  \\ 
\hline 
 & \multicolumn{21}{c}{\cellcolor{gray}{\textcolor{white}{\it{ImageNet}}}}   \\ \cline{2-22}
 
   Narrow & \bf{11.15} & \underline{0.071} & \bf{0.812} &12.81 &0.124 &0.762 & 12.74 & 0.116 & 0.769& 12.69 &0.125 &0.768 &\underline{11.75} &0.083 &0.793 &11.97 &\bf{0.068} & \underline{0.804}&12.85 &0.126 &0.760  \\
 
  Wide1 & \bf{13.24} & \bf{0.121} & \bf{0.785} &14.28 &0.141 &0.756 & 14.16 & 0.137 & 0.760&14.21 &0.135 &0.754 & 13.98 &\underline{0.125} &\underline{0.781} &\underline{13.93} & 0.132 &0.772 &14.25 &0.138 &0.758  \\
  
  Wide2 & \bf{13.37} & \underline{0.134} & \bf{0.776} & 14.42&0.157 &0.739 & 14.35 & 0.151 & 0.749&14.46 &0.158 &0.741 & 14.27 & 0.146 & \underline{0.764} &\underline{14.11} &\bf{0.129} &0.759 &14.40 &0.154 &0.741  \\
 
  Half ver. & \bf{15.59} & \bf{0.283} & \bf{0.614} &22.67 &0.338 &0.562 & 21.63 & 0.308 & 0.594&22.47 &0.322 &0.578 & {17.43} &\underline{0.292} &\underline{0.604} & \underline{17.38} & 0.298 & 0.598 &22.80 &0.342 &0.553 \\
  
  Half hor. & \bf{15.24} & \bf{0.271} & \bf{0.622} &22.53 &0.324 &0.582 & 21.84 & 0.302 & 0.603&22.41 &0.317 &0.594 &{17.36} &0.289 & \underline{0.607} &\underline{17.29} &\underline{0.284} &\underline{0.607} &22.61 &0.330 & 0.570  \\
 
 {Expand} & \bf{50.35} & \bf{0.574} & \bf{0.349} &63.21 &0.674 &0.271 & 60.86 & 0.627 & 0.307&59.97 &\underline{0.605} & \underline{0.328} &62.85 &0.659 &0.298 & \underline{59.43} & 0.618 &{0.314} &63.15 &0.670 & 0.277 \\
\hline  
{Avg } & \bf{19.83} & \bf{0.237} & \bf{0.660} & 24.99 &0.293 &0.612 & 24.26 & 0.274 & 0.630 & 24.36 & 0.277 & 0.625 & 22.94 & 0.266 & 0.640 & \underline{22.35} & \underline{0.254} & \underline{0.643} & 25.02 & 0.295 & 0.610  \\
\hline
\end{tabular}
}
\end{table*}
\begin{figure}
\centering
\includegraphics[width=0.99\columnwidth]{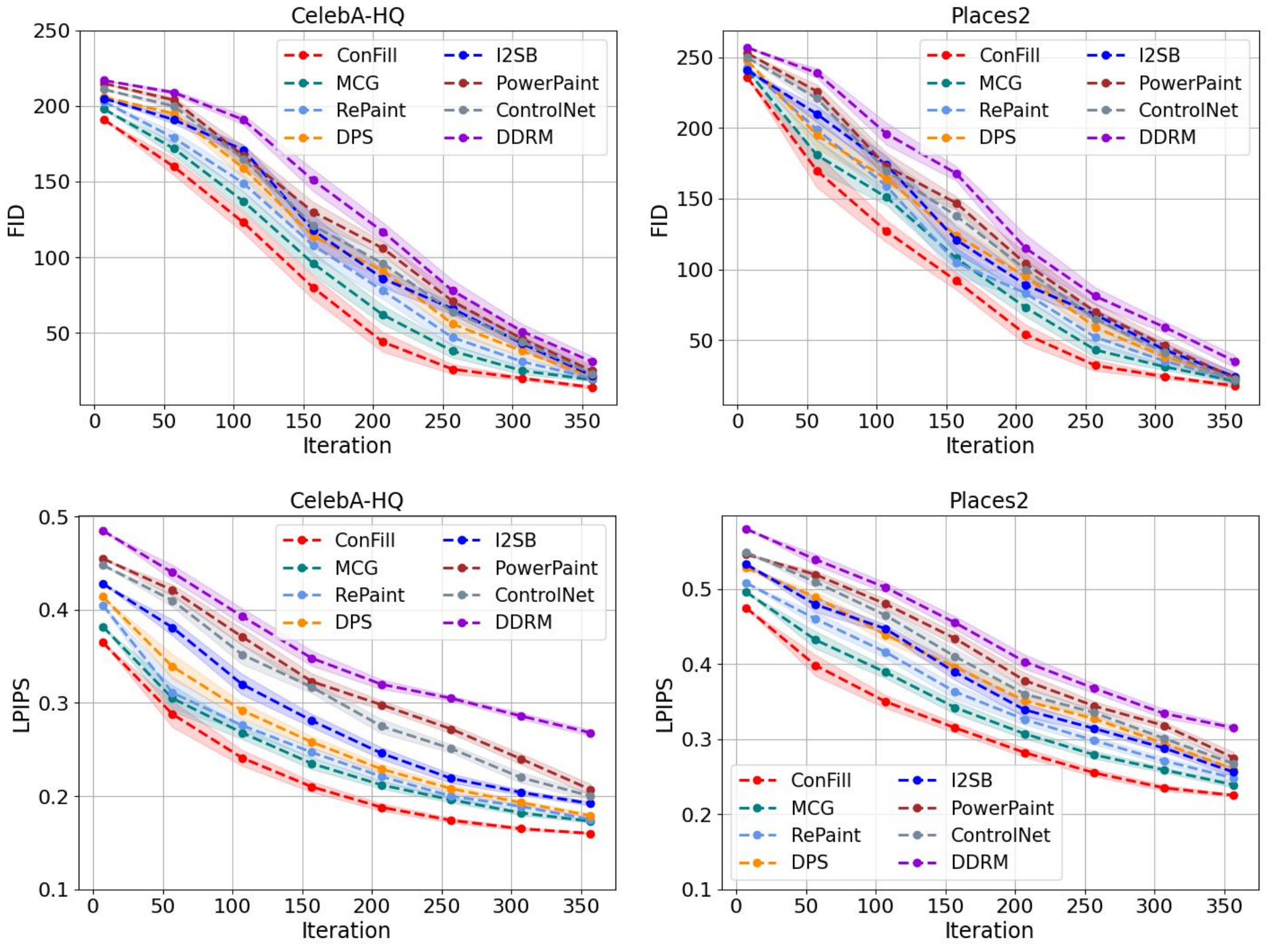} 
\vspace{-3pt}
\caption{\small{Evaluation of FID (top row) and LPIPS (bottom row) scores for ConFill and baseline models on the CelebA-HQ and Places2 datasets. ConFill consistently outperforms the other models in both fidelity and perceptual similarity at all iteration levels.}}  
\label{FID}
\end{figure}
\noindent 
We have observed that directly optimising $\hat{x}_{t}$ using the gradient often leads to NaN values. To mitigate this issue, following \cite{zhang2023towards} we fine-tune the learning rate by introducing an additional term to optimise the training process more effectively.
Therefore, we start with an initial learning rate of $0.01$ and then adjust it using an adaptive learning rate to fine-tune our learning process. We initially set $\upsilon =0.1$ and $\tau = 0.02$.
Moreover, ConFill's uses a travel frequency of \(\mathcal{J} = 1\) and a time travel interval of \(\acute{\mathfrak{t}} = 10\).
\subsection{Comparison to other Established Methods}
This section presents a detailed comparison between ConFill and recent diffusion-based image completion methods.
\vspace{2pt}
\begin{figure*}
\centering
\includegraphics[width=1.95\columnwidth]{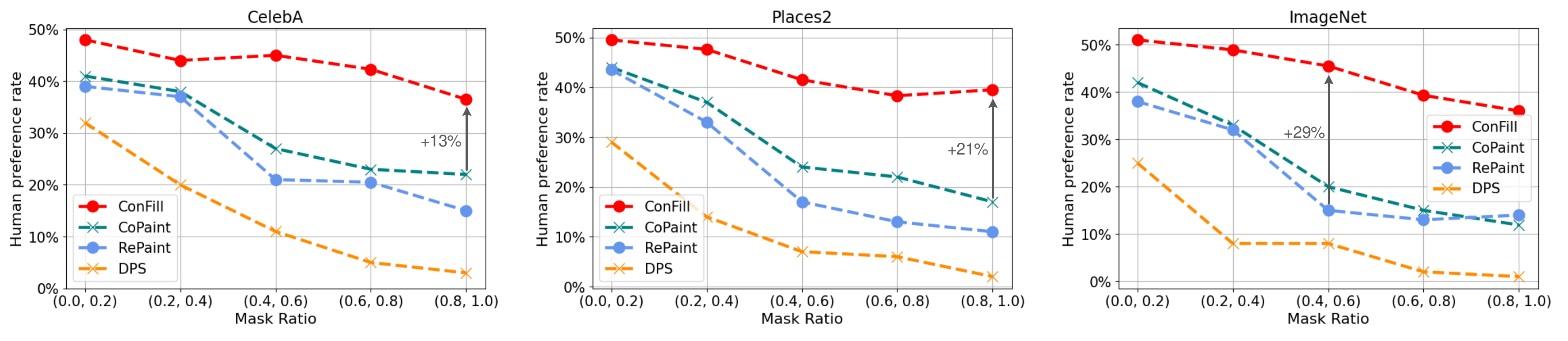}
\vspace{-4pt}
\caption{\small{The results of the user study comparing our model, ConFill, with baseline models CoPaint, RePaint, and DPS across various mask ratios on three different datasets: CelebA, Places2, and ImageNet. ConFill consistently outperforms the other models across all datasets, as demonstrated by the higher human preference rates, especially at lower mask ratios. Notably, ConFill achieves a +21\% increase on the Places2 dataset and a +29\% increase on the ImageNet dataset compared to the best-performing baseline models.}}  
\label{user}
\end{figure*}
\begin{figure*}
\centering
\includegraphics[width=1.52\columnwidth]{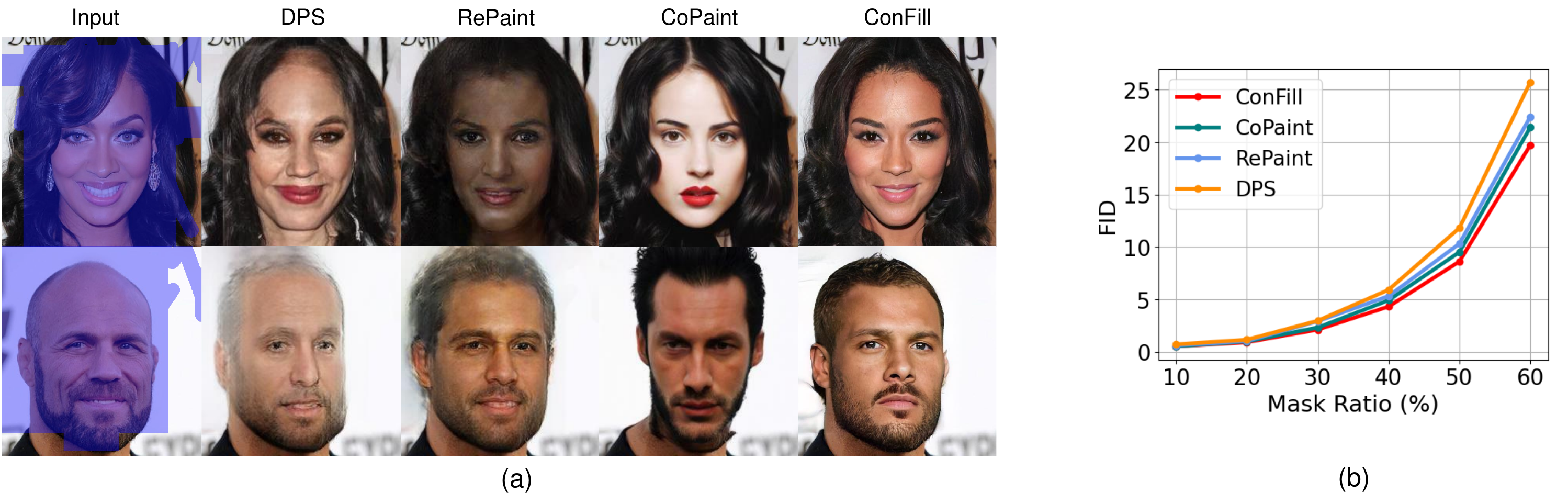} 
\vspace{-5pt}
\caption{\small{A comparison of image inpainting results on the facial images with large unseen holes (left) and a corresponding evaluation of FID scores with varying mask ratios (right). ConFill demonstrates superior performance in generating realistic and coherent facial features, with smoother integration and fewer artifacts than DPS, RePaint and CoPaint. As the mask ratio increases, ConFill consistently achieves lower FID scores, indicating its effectiveness in preserving image quality under increasingly challenging conditions.}}
\label{l-mask}
\end{figure*}

\noindent{\bf Qualitative Comparisons:} 
In Figs. 3, 4, and 5, we show sample images generated using datasets from CelebA-HQ, Places2, and ImageNet-1k, respectively. Two key observations can be drawn from these results. Firstly, ConFill generates more realistic content with better coherence and semantic consistency compared to baseline methods, particularly for larger masks like Expand and Half. For example, in the fifth row of Fig. \ref{place}, where only half of a desk is visible in the masked region, most baseline methods fail to produce satisfactory completions. In contrast, ConFill successfully reconstructs a desk that aligns with the input image in both size and style. Additionally, while some approaches, like DPS and CoPaint, do create relatively realistic images, our method consistently generates images with greater consistency. As an example, in the last row of Fig. \ref{place}, the sunset view produced by our method shows greater consistency with the input image, whereas the versions produced by DPS and CoPaint are less detailed. Our model demonstrates robust performance, enabling the generation of high-quality outputs that remain coherent with the existing image content. This efficacy is attributed to our model’s innovative refining posterior distribution estimation mechanism during denoising. This mechanism starts with the visible pixels and methodically diffuses essential information throughout the image, assuring flawless integration of the generated content with the known regions of the image. This approach is particularly effective for facial images that show strong inherent characteristics like symmetry and structural features. For example, in the fourth line of Fig. \ref{celeba}, where the bottom half of the face is masked, our ConFill successfully uses visible features to generate a realistic and consistent result. This capacity is also demonstrated in natural scenes, for instance, in the fourth example in Fig. \ref{imagenet}, our method accurately restores the structure of a flower, outperforming other methods. Overall, these results indicate the effectiveness of our approach across a variety of contexts. Our framework reconstructs fine-grained textures and edges (local information) while maintaining overall semantic and structural coherence (global information), even in regions with large occlusions.

\noindent{\bf Quantitative Comparisons:}
Table \ref{tab.1} reports the quantitative results of ConFill, as well as various baseline methods, across CelebA-HQ, Places2, and ImageNet-1k datasets using different mask types. Here are the notable observations derived from our analysis. Firstly, concerning the evaluation metrics, ConFill consistently surpasses other baseline methods, demonstrating an average decrease in FID score by $2.1\%$, $2.62\%$, and $2.53\%$ compared to the top-performing baseline, CoPaint, across the CelebA-HQ, Places2, and ImageNet-1k datasets, respectively. Furthermore, the images generated by ConFill show improved coherence and outperform in several other respects, including naturalness and complying with the image completion constraint. Moreover, it is important to observe that ConFill performance advantage is particularly notable on the Places2 and ImageNet-1k datasets. This is likely because these datasets contain more complex images, making any imperfections, including incoherence, more obvious.
For example, on Places2 with an expand mask, ConFill has $3.4\%$ lower FID score compared to CoPaint. On the other hand, on ImageNet-1k our model has an average LPIPS score of $0.242$, which is $4.8\%$ and $8.6\%$ higher than CoPaint and RePaint, respectively. As demonstrated in Figs. \ref{place} and \ref{imagenet}, ControlNet and PowerPaint struggle when synthesising images with large holes, whereas ConFill produces realistic images with fewer artefacts than other methods. 

Fig. \ref{FID} highlights the superior performance of ConFill compared to the state-of-the-art models.  
ConFill converges faster than baseline models due to CAD, which efficiently guides the diffusion process to produce high-quality samples that align closely with the target distribution in fewer iterations. Notably, after just 50k iterations, ConFill achieves a 62-point improvement in FID on Places2, demonstrating its effectiveness in enhancing image quality and perceptual similarity with greater efficiency.


\begin{figure}
\centering
\includegraphics[width=0.97\columnwidth]{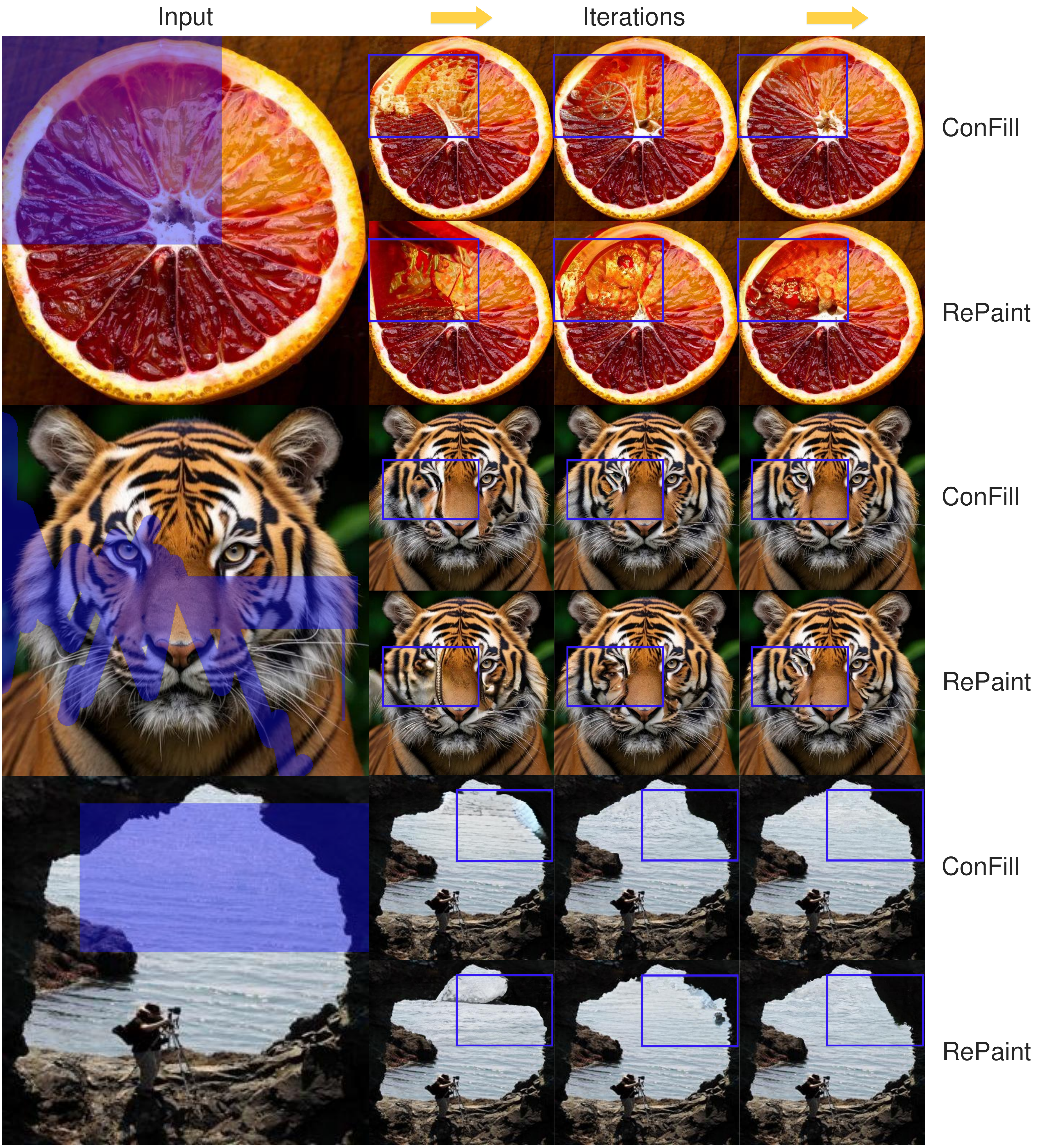} 
\vspace{-4pt}
\caption {\small{The image completion results of our ConFill and RePaint across various iterations. Initially, both models generate artifact-laden outputs. However, as iterations progress, ConFill consistently produces sharper and more accurate reconstructions compared to RePaint. For example, our model produces a more realistic reconstruction of the missing fruit segment, with improved color and texture, while RePaint retains artifacts and shows less consistent color matching.}}
\label{iterative}
\end{figure}
\vspace{-3pt}
\subsection{Human Perceptual}
We conducted a subjective analysis through a human evaluation with different mask ratios. Participants are shown a reference image with masked parts, alongside two completed versions: one by ConFill and the other by a baseline method. They chose the higher-quality image based on predefined criteria, focusing on the coherence of the results. Feedback from 46 participants was collected, evaluating 2,240 image pairs from CelebA-HQ, Places2, and ImageNet datasets. The vote difference, calculated as a percentage, indicates the preference for ConFill over baseline models. In Fig. \ref{user}, ConFill was 21\% more likely to be selected over CoPaint on Places2 with a mask ratio of 0.8-1.0 and 29\% more likely over RePaint on ImageNet with a mask ratio of 0.4-0.6.

\vspace{-3pt}
\subsection{Adaptation to Unseen Mask Types}
Based on the guidelines of the NTIRE 2022 Image Inpainting Challenge \cite{romero2022ntire}, we developed a test set comprising $5,000$ samples, each with an average missing ratio of approximately $60\%$. This set includes mask types such as Image Expansion and Nearest Neighbour, which were not included during the training phase. As illustrated in Table \ref{tab.4}, other state-of-the-art models, particularly DPS, RePaint and CoPaint show performance declines. In contrast, our method shows excellent generalisation ability for these previously unseen mask types. Our method achieved a superior FID score of 19.63, representing a 17.8\% improvement over the second-best approach, CoPaint, which scored 23.87. This success is due to our CAD-based denoising mechanism, which iteratively refines the model’s predictions, leading to more accurate image completion. The percentage improvement is calculated by \((\frac{{\text{Baseline performance}}-{\text{new performance}}}{\text{Baseline performance}}) \times 100\). 
%
\begin{table}
\centering
\footnotesize{
\caption{\small{Quantitative analysis and inpainting speed on previously unseen mask types. ConFill achieves a 17.8\% improvement in FID, 12.2\% in LPIPS, and 4.0\% in SSIM compared to the closest competitor, CoPaint. }}
\vspace{-4pt}
\label{tab.4}
\begin{tabular}{l|c c c c}     \hline
\rule{0pt}{1\normalbaselineskip}Method & FID$\downarrow$ & LPIPS$\downarrow$&  {SSIM}$\uparrow$ &  {Runtime}$\downarrow$ \\  [0.2ex]  \hline 
\rowcolor{maroon!15}
\rule{0pt}{1\normalbaselineskip}ConFill (ours) & {\bf{19.63}} & {\bf{0.173}} & {\bf{0.752}}& 109.24 \\ \hline 
\rule{0pt}{1\normalbaselineskip}CoPaint \cite{zhang2023towards}  & 23.87  &  0.197& 0.723 & 117.96  \\ 

PowerPaint \cite{zhuang2025task}   & 27.46  &0.221  &0.704 & 274.83    \\ 

RePaint \cite{lugmayr2022repaint}  & 25.32  &  0.213& 0.708&  210.58  \\ 

ControlNet \cite{zhang2023adding}  & 27.83  &0.235  &0.697  & 223.26  \\ 

DPS \cite{chung2022diffusion}  & 28.76  &  0.238 & 0.685& 453.05 \\  
{I$^{2}$SB} \cite{liu20232}  & 29.25  &  0.241 & 0.680& 178.67 \\
DDRM \cite{kawar2022denoising}  & 43.51  &  0.306 & 0.629& 162.41 \\ 
\hline
\end{tabular}}
 \end{table}
Additionally, we evaluate time efficiency by calculating the average runtime required to sample an image. The results indicate that ConFill achieves faster sampling compared to other methods. The longer sampling times in RePaint is due to its resampling mechanism that needed to synchronize condition and generation, while CoPaint requires an additional computation due to its extra backward pass and the need for more optimization steps. 
ControlNet introduces conditional control to diffusion, adding moderate computational overhead, while PowePaint, relies on task prompts, making it more time-intensive. In contrast, ConFill reduces inference time by using an optimized reverse diffusion schedule and a lightweight denoising architecture, which minimizes computational overhead without compromising image quality.

Fig. \ref{l-mask} (left) demonstrates that the baseline models struggle to accurately reconstruct images with extremely large unseen masks, leading to outputs with notable distortions. In contrast, ConFill consistently produces reconstructions of superior quality. Additionally, Fig. \ref{l-mask} (right) illustrates the relationship between FID scores and the mask ratio. While RePaint and COPaint show FID scores comparable to our model at the minimal mask ratio, ConFill surpasses them by a greater margin as the mask size increases, indicating its effectiveness for unseen, larger missing regions.
\begin{figure}
\centering
\includegraphics[width=0.97\columnwidth]{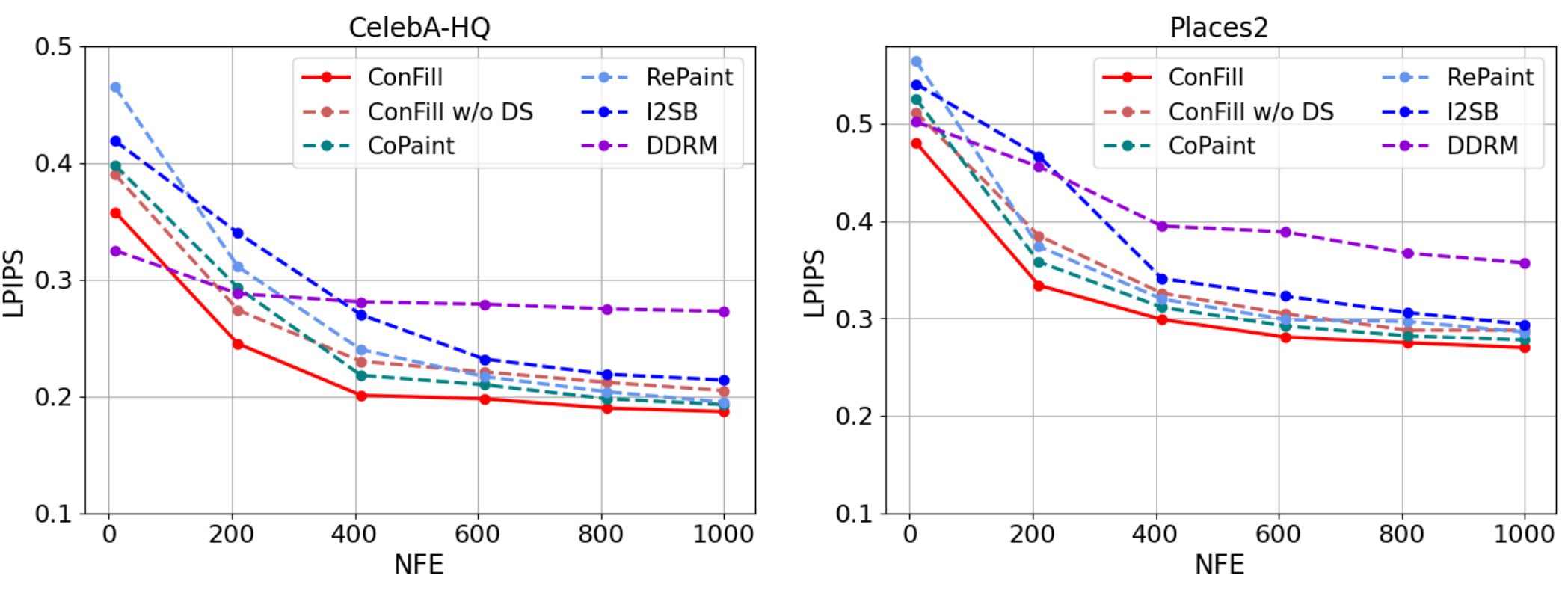} 
\vspace{-4pt}
\caption{\small{Ablation analysis using a half-mask on the CelebA and Places2 datasets. Shows the performance of ConFill with and without dynamic sampling (DS) and compares it with the other models. Results show ConFill with DS performs strongly even at lower NFEs. }}  
\label{nfe}
\end{figure}

\vspace{-3pt}
\subsection{Iterative Performance Analysis}
Fig. \ref{iterative} shows the incremental results across different iterations, indicating that our method reaches promising performance with fewer iterations, demonstrating a faster performance than RePaint. This suggests that our model is not only faster but also more efficient in terms of iteration count needed to reach a refined result, demonstrating its potential for applications that require high-quality image restoration with fewer computational resources.

\begin{figure*}
\centering
\includegraphics[width=1.84\columnwidth]{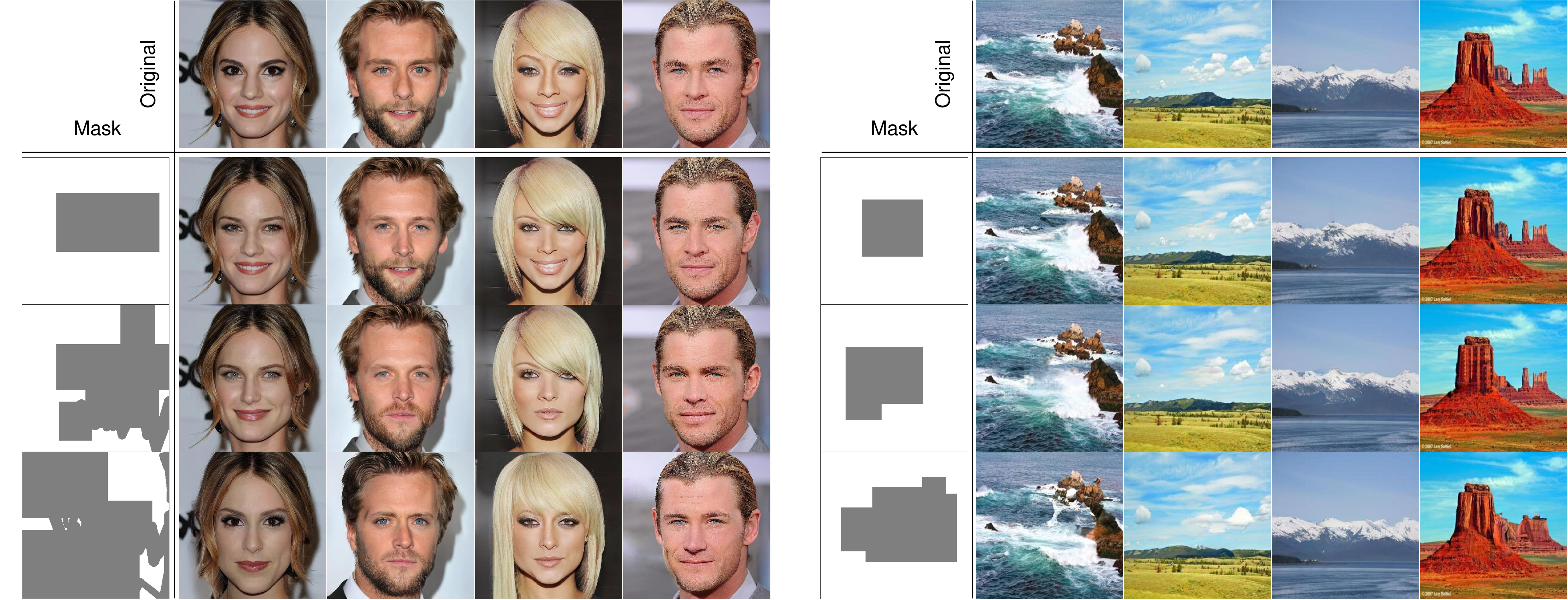} 
\vspace{-4pt}
\caption{\small{Image completion results of ConFill for the CelebA and Places2 datasets using various mask types (small to large size). }}  
\label{1-img}
\end{figure*}

\setlength\dashlinedash{0.4pt}
\setlength\dashlinegap{1.5pt}
\setlength\arrayrulewidth{0.3pt}
\begin{table}[t]
\centering
\footnotesize{
\caption{\small{Ablation Analysis. Performance evaluation of our model using CAD, standard WD and Euclidean distance with varying gradient descent steps ($G$). In this evaluation, we use the test set of Places2 with wide mask. The best settings are highlighted.}}
\vspace{-5pt}
\label{ablation}
\begin{tabular}{l c| c c c}    \hline
 Method & G & FID$\downarrow$ & LPIPS$\downarrow$&  {SSIM}$\uparrow$  \\  \hline

\rowcolor{maroon!15}
\cellcolor{maroon!15}  & 1  & 10.48  &  0.129& 0.808  \\

\rowcolor{maroon!30}
\cellcolor{maroon!15} ConFill w CAD (ours) & \bf2  & \bf10.39  & \bf0.120&  \bf0.812   \\ 

\rowcolor{maroon!15}
\cellcolor{maroon!15}  &5  & 10.56  &  0.133& 0.801   \\  \hdashline 

&1 & 12.35 & 0.152 & 0.779 \\ 
ConFill w WD &2  & 11.97  &  0.141& 0.791   \\  
 &5  & 12.43  &  0.154& 0.769   \\  \hdashline 

 &1 & 12.28 & 0.154 & 0.784 \\ 
ConFill w $L2$ &2  & 12.32  &  0.150& 0.784   \\  
 &5  & 12.61  &  0.166& 0.759   \\  \hline

\end{tabular}}
 \end{table}

\begin{table}
\centering
\footnotesize{
\caption{\small{Impact of time travel interval ($\acute{\mathfrak{t}}$) and frequency ($\mathcal{J}$). The time travel interval of 10 and a frequency of 1 provide the most effective balance between image quality and structural similarity.}}
\vspace{-5pt}
\label{ablation-1}
\begin{tabular}{l c| c c c}    \hline

 Method & $\acute{\mathfrak{t}}$ & FID$\downarrow$ & LPIPS$\downarrow$&  {SSIM}$\uparrow$  \\    \hline

 &2 & 10.48 & 0.126 & 0.807 \\ 

 ConFill &5  & 10.44  &  0.123& 0.809   \\

  &\cellcolor{maroon!15} \bf10  & \cellcolor{maroon!15} \bf10.39  & \cellcolor{maroon!15} \bf 0.120& \cellcolor{maroon!15}\bf 0.812  \\ 

 &15  & 10.41  &  0.119& 0.810  \\  \hline 

  & $\mathcal{J}$ & FID$\downarrow$ & LPIPS$\downarrow$&  {SSIM}$\uparrow$  \\    \hline
 &\cellcolor{maroon!15} \bf1 & \cellcolor{maroon!15} \bf10.39 & \cellcolor{maroon!15} \bf0.120 & \cellcolor{maroon!15} \bf0.812 \\ 

 ConFill &2  & 10.40  &  0.121& 0.810   \\ 

 &5  & 10.42  &  0.122& 0.809    \\  \hline
\end{tabular}}
 \end{table}

\vspace{-6pt}
\subsection{Ablation Analysis}
We investigate the contribution of CAD to our model by comparing the performance of our ConFill (which uses CAD), with a variant that uses standard WD and $L2$. We also examine the effects of three hyperparameters: gradient descent steps (G), time travel interval ($\acute{\mathfrak{t}}$) and its frequency ($\mathcal{J}$). Our experiments are conducted on Places2 with wide mask and the results are detailed in Table \ref{ablation} and Table \ref{ablation-1}.
Our model uses \(G\)-step gradient descent approach to optimize \(\hat{x}_t\) at every timestep. However, as shown in Table \ref{ablation}, increasing \(G\) does not improve performance or make the generated images better match the real ones. Because the optimization of \(\hat{x}_t\) focuses on minimising the mean square error in the known regions (not masked), as detailed in the second term of Eq. (\ref{eq:17}). CAD inherently provides smoother and more meaningful gradients during optimization. This makes each gradient descent step more effective at reducing the gap between generated images and real ones while also adapting to local contextual differences. 
However, a higher number of gradient steps can lead to overfitting. 
Moreover, the analysis shows that ConFill with WD provides more stable and consistent performance across all gradient steps compared to ConFill with $L2$. While $L2$ performs slightly better at the initial step (\(G = 1\)), its performance degrades as $G$ increases, whereas WD maintains higher stability. 
\begin{table}
\centering
\footnotesize{
\caption{\small {Performance comparison on different dynamic sampling configurations. Evaluation on CelebA with an expanded mask.}}
\vspace{-5pt}
\label{tab.8}
\setlength{\tabcolsep}{4.5pt} 
\begin{tabular}{l|c c c}     \hline
\rule{0pt}{1\normalbaselineskip}Implementation & FID$\downarrow$ & LPIPS$\downarrow$&  {SSIM}$\uparrow$  \\  [0.2ex]  \hline 
\rowcolor{maroon!15}
\rule{0pt}{1\normalbaselineskip}Combined sampling & {\bf{14.27}} (+5.2\%) & {\bf{0.164}} (+6.9\%) & {\bf{0.760}} (+6.3\%) \\ \hline 
\rule{0pt}{1\normalbaselineskip}Texture-driven sampling   & 15.86  &  0.178& 0.753    \\  
Edge-driven sampling  & 16.21  &  {0.182}& 0.748    \\  

\hline
\end{tabular}}
 \end{table}

\begin{table}
\centering
\footnotesize{
\caption{\small{The impact of $ \upsilon$, which controls the scaling of the distance function based on the contextual information and $\tau$ affects sensitivity to contextual differences, is evaluated using the ImageNet test set with a wide mask. The best settings are highlighted, with improvements (\%) shown between the most and least effective settings.}}
\label{ablation-2}
\begin{tabular}{@{}l c|c c c@{}}    \hline
 Method &  $ \upsilon$ & FID$\downarrow$ & LPIPS$\downarrow$&  {SSIM}$\uparrow$  \\    \hline 
 
&~0.05 & ~13.43  &  ~0.130& ~0.764  \\
ConFill & \cellcolor{maroon!15} {\bf 0.10}  & \cellcolor{maroon!15} {\bf 13.24} (+1.4\%)  & \cellcolor{maroon!15} {\bf 0.121} (+6.9\%) & \cellcolor{maroon!15} {\bf 0.785} (+2.7\%)  \\ 
&~0.15  & ~13.26  &  ~0.122& ~0.783   \\  \hline
 & $\tau$ & FID$\downarrow$ & LPIPS$\downarrow$&  {SSIM}$\uparrow$  \\    \hline 


  &~0.01  &~13.26  & ~0.121 & ~0.786   \\  

ConFill  &\cellcolor{maroon!15} \bf 0.02  &\cellcolor{maroon!15} \bf 13.24 & \cellcolor{maroon!15} \bf 0.121 & \cellcolor{maroon!15} \bf 0.785  \\ 

  &~0.03  & ~13.27  &  ~0.120& ~0.784  \\ 

\hline
\end{tabular}}
 \end{table}

In Table \ref{ablation-1}, we evaluate the impact of the time travel interval \(\acute{\mathfrak{t}}\) and frequency \(\mathcal{J}\) on model performance.  The results show that \(\acute{\mathfrak{t}}\) has minimal impact when \(\acute{\mathfrak{t}} \geq 5\). Additionally, unlike RePaint, which uses a time travel frequency of \(\mathcal{J}=9\), our approach performs well with \(\mathcal{J}=1\). This suggests that our model more effectively enforces image completion constraints compared to the simple replacement techniques used in RePaint. 
To evaluate the role of different criteria for dynamic sampling, we analyze different sampling configurations: I) Texture-driven sampling ($\hat{\alpha}=1$, $\hat{\beta}=0$) where sampling weights are only based on local intensity variance. II) Edge-driven sampling ($\hat{\alpha}=0$, $\hat{\beta}=1$) where sampling weights are only based on edge density to focus on regions with prominent spatial structures. III) Combined sampling ($\hat{\alpha}=0.5$, $\hat{\beta}=0.5$) where sampling weights are computed as a balanced combination of both measures. Table \ref{tab.8} summarizes the results for each configuration. The combined approach improves reconstruction quality, particularly in regions with complex textures.

In Table \ref{ablation-2} we evaluate the impact of the parameters $\upsilon$ and $\tau$. The results indicate that $\upsilon$ has a significant effect on FID and LPIPS as it directly scales the distance metric. For instance, adjusting \(\upsilon\) from 0.05 to 0.10 results in a 1.4\% improvement in FID (from 13.43 to 13.24) and a 6.9\% improvement in LPIPS (from 0.130 to 0.121). However, values beyond 0.1 do not result in further notable changes. Meanwhile, \(\tau\) controls the sensitivity of the scaling function to contextual differences. 
Fig. \ref{nfe} compares the performance of different methods in terms of LPIPS across varying numbers of function evaluations (NFE) on CelebA-HQ and Places2. ConFill consistently achieves the lowest LPIPS values across all NFEs, demonstrating its effectiveness in minimizing perceived differences regardless of the NFE. ConFill without dynamic sampling (DS) maintains low LPIPS values but is slightly less effective compared to ConFill with DS. RePaint starts with a relatively high LPIPS value but shows significant improvement as NFEs increase. Despite this improvement, RePaint still falls below ConFill.
To analyze the impact of \(C(\cdot)\) in the image completion process, we evaluate its performance under different configurations. Handcrafted features only, learned features only, and a combination of handcrafted and learned features. Table \ref{tab.7} reports the results, demonstrating that combining handcrafted and learned features consistently outperforms the individual ones in terms of perceptual quality and contextual coherence.
The results of our model using various mask types are shown in Fig. \ref{1-img}. ConFill connects different conditioned instances, ranging from small to large completions and generates plausible and coherent content for the missing areas.

\begin{table}
\centering
\footnotesize{
\caption{\small {Performance comparison of contextual feature extractor implementations. Evaluation on ImageNet with an expanded mask.}}
\vspace{-5pt}
\label{tab.7}
\setlength{\tabcolsep}{4.5pt} 
\begin{tabular}{l|c c c}     \hline
\rule{0pt}{1\normalbaselineskip}Implementation & FID$\downarrow$ & LPIPS$\downarrow$&  {SSIM}$\uparrow$  \\  [0.2ex]  \hline 
\rowcolor{maroon!15}
\rule{0pt}{1\normalbaselineskip}Combined features & {\bf{50.35}} (+5.2\%) & {\bf{0.574}} (+6.9\%) & {\bf{0.349}} (+6.3\%) \\ \hline 
\rule{0pt}{1\normalbaselineskip}Handcrafted features   & 55.34  &  0.657& 0.308    \\  
Learned features  & 52.95  &  {0.614}& 0.327    \\  

\hline
\end{tabular}}
 \end{table}

\setlength\dashlinedash{0.4pt}
\setlength\dashlinegap{1.5pt}
\setlength\arrayrulewidth{0.3pt}
\begin{table}[t]
\centering
\footnotesize{
\caption{\small {Impact of different sample size on FID, LPIPS, and SSIM metrics for different methods. Evaluation on Places2 with an expanded mask.}}
\vspace{-5pt}
\label{sample size}
\begin{tabular}{l c| c c c}    \hline
 Sample size & Method & FID$\downarrow$ & LPIPS$\downarrow$&  {SSIM}$\uparrow$  \\  \hline

  & RePaint  & 58.72   & 0.614  & 0.357   \\

500 images &  PowerPaint  & 56.90   & 0.598 & 0.371     \\ 

  & CoPaint  & 56.37   & 0.588  & 0.381    \\  
  
\rowcolor{maroon!15}
\cellcolor{white}  & ConFill  & 47.12   & 0.540  & 0.419    \\  \hdashline 

  & RePaint  & 58.21   & 0.612  & 0.359   \\ 

1000 images & PowerPaint  & 56.34   & 0.599  & 0.373    \\  

  & CoPaint  & 56.05   & 0.585  & 0.383    \\  
 
\rowcolor{maroon!15}
\cellcolor{white}  & ConFill  & 46.58   & 0.540  & 0.420    \\ \hline 

\end{tabular}}
 \end{table}

In Table \ref{sample size}, we analyse the performance of different models as the evaluation set size increases. Across all metrics, we observe consistent trends. FID decreases with larger evaluation sets, reflecting improved alignment with the data distribution. LPIPS and SSIM remain relatively stable, with minor improvements, indicating the reliability and robustness of the metrics across varying sample sizes. These trends demonstrate that the evaluation protocol effectively captures the performance of the models, providing a comprehensive understanding of their behavior as sample sizes increase.
\begin{table}[h!]
\centering
\footnotesize{
\caption{\small{Quantitative analysis with non-diffusion models on Places2 with an expanded mask. Compared to MAT, ConFill shows a 3.4\% improvement in FID, and 1.8\% in SSIM.}}
\vspace{-5pt}
\label{tab.5}
\begin{tabular}{l|c c c}     \hline
\rule{0pt}{1\normalbaselineskip}Method & FID$\downarrow$ & LPIPS$\downarrow$&  {SSIM}$\uparrow$  \\  [0.2ex]  \hline 
\rowcolor{maroon!15}
\rule{0pt}{1\normalbaselineskip}ConFill (ours) & {\bf{48.65}} (+3.4\%) & 0.543 & {\bf{0.417}} (+1.8\%) \\ \hline 
\rule{0pt}{1\normalbaselineskip}LaMa \cite{suvorov2022resolution}   & 55.46  &  0.558& 0.398    \\  
MAT \cite{li2022mat}   & 50.31  &  \bf{0.540}& 0.410    \\  

\hline
\end{tabular}}
 \end{table}

\subsection{Comparison to Non-diffusion Models}
To evaluate the performance of ConFill against non-diffusion models, we selected two state-of-the-art methods: LaMa \cite{suvorov2022resolution} and MAT \cite{li2022mat}, which are benchmarks in GAN and Transformer-based architectures, respectively. By comparing ConFill with these models, we aim to show its strengths, particularly in visual quality and coherence, as shown in Table \ref{tab.5}. Specifically, ConFill achieves an FID score of 50.35, outperforming LaMa (56.46) and MAT (52.31), demonstrating a stronger alignment with the original image distribution. Furthermore, in terms of LPIPS, MAT with a small margin performs better than ConFill which is due to its specific training with large masks. However, ConFill achieves a higher SSIM score of 0.417, outperforming the other models. These results highlight ConFill's robustness in generating visually coherent and high-quality images, especially for large masked regions where maintaining visual coherence is particularly challenging. This is achieved by integrating context-adaptive discrepancy, which effectively balances the trade-offs between reconstruction fidelity and perceptual quality.

\section{Conclusion}
In this paper, we proposed ConFill, a novel approach that adopts the principled application of CAD within a diffusion model framework to address the challenges of image completion tasks. ConFill introduces a novel strategy for generative inverse problems by effectively minimizing the CAD between the latent representations of noisy and original data distributions. This refined approach ensures smooth integration of reconstructed regions with their surrounding context and dynamically adapts to local image features.
By iteratively refining the posterior distribution estimation during the denoising process, ConFill substantially reduces errors and imposes strict penalties on discrepancies with the reference image. The integration of CAD enhances this process, allowing the model to adjust its computations based on the contextual differences between image segments, significantly improving the fidelity and coherence of the completed images. Additionally, ConFill incorporates a dynamic sampling strategy that adjusts the sampling process based on contextual information, further improving the efficiency and quality of image completion. Our extensive evaluations on three datasets have demonstrated ConFill's superiority over current state-of-the-art methods.

\section*{Acknowledgment}
This project is partially supported by the University of York Research Grant No. M0259102, and the National Research Foundation, Singapore, under its NRF Professorship Award No. NRF-P2024-001. Also supported by the Swedish Research Council (2022-04266) and KAW (DarkTree project; 2024.0076).

{\footnotesize{
\bibliographystyle{IEEEtran}
\bibliography{IEEEabrv,IEEEexample}

\begin{thebibliography}{10}
\providecommand{\url}[1]{#1}
\csname url@samestyle\endcsname
\providecommand{\newblock}{\relax}
\providecommand{\bibinfo}[2]{#2}
\providecommand{\BIBentrySTDinterwordspacing}{\spaceskip=0pt\relax}
\providecommand{\BIBentryALTinterwordstretchfactor}{4}
\providecommand{\BIBentryALTinterwordspacing}{\spaceskip=\fontdimen2\font plus
\BIBentryALTinterwordstretchfactor\fontdimen3\font minus \fontdimen4\font\relax}
\providecommand{\BIBforeignlanguage}[2]{{%
\expandafter\ifx\csname l@#1\endcsname\relax
\typeout{** WARNING: IEEEtran.bst: No hyphenation pattern has been}%
\typeout{** loaded for the language `#1'. Using the pattern for}%
\typeout{** the default language instead.}%
\else
\language=\csname l@#1\endcsname
\fi
#2}}
\providecommand{\BIBdecl}{\relax}
\BIBdecl

\bibitem{ling2021editgan}
H.~Ling, K.~Kreis, D.~Li, S.~W. Kim, A.~Torralba, and S.~Fidler, ``Editgan: High-precision semantic image editing,'' \emph{Proc. Advances Neural Inf. Process. Syst.}, vol.~34, pp. 16\,331--16\,345, 2021.

\bibitem{li2022sdm}
W.~Li, X.~Yu, K.~Zhou, Y.~Song, Z.~Lin, and J.~Jia, ``Image inpainting via iteratively decoupled probabilistic modeling,'' \emph{arXiv preprint arXiv:2212.02963}, 2022.

\bibitem{he2019adversarial}
R.~He, J.~Cao, L.~Song, Z.~Sun, and T.~Tan, ``Adversarial cross-spectral face completion for nir-vis face recognition,'' \emph{IEEE Trans. Pattern Anal. Mach. Intell.}, vol.~42, no.~5, pp. 1025--1037, 2020.

\bibitem{li2022all}
B.~Li, X.~Liu, P.~Hu, Z.~Wu, J.~Lv, and X.~Peng, ``All-in-one image restoration for unknown corruption,'' in \emph{Proc. Conf. Comput. Vis. Pattern Recognit.}, 2022, pp. 17\,452--17\,462.

\bibitem{yildirim2023inst}
A.~B. Yildirim, V.~Baday, E.~Erdem, A.~Erdem, and A.~Dundar, ``Inst-inpaint: Instructing to remove objects with diffusion models,'' \emph{arXiv preprint arXiv:2304.03246}, 2023.

\bibitem{suvorov2022resolution}
R.~Suvorov, E.~Logacheva, A.~Mashikhin, A.~Remizova, A.~Ashukha, A.~Silvestrov, N.~Kong, H.~Goka, K.~Park, and V.~Lempitsky, ``Resolution-robust large mask inpainting with fourier convolutions,'' in \emph{Proc. Winter Conf. Appl. Comput. Vis.}, 2022, pp. 2149--2159.

\bibitem{lu2022glama}
Z.~Lu, J.~Jiang, J.~Huang, G.~Wu, and X.~Liu, ``Glama: Joint spatial and frequency loss for general image inpainting,'' in \emph{Proc. Conf. Comput. Vis. Pattern Recognit. Workshop}, 2022, pp. 1301--1310.

\bibitem{zeng2022aggregated}
Y.~Zeng, J.~Fu, H.~Chao, and B.~Guo, ``Aggregated contextual transformations for high-resolution image inpainting,'' \emph{IEEE Trans. Vis. Comput. Graph.}, 2022.

\bibitem{li2022mat}
W.~Li, Z.~Lin, K.~Zhou, L.~Qi, Y.~Wang, and J.~Jia, ``Mat: Mask-aware transformer for large hole image inpainting,'' in \emph{Proc. Conf. Comput. Vis. Pattern Recognit.}, 2022, pp. 10\,758--10\,768.

\bibitem{shamsolmoali2023transinpaint}
P.~Shamsolmoali, M.~Zareapoor, and E.~Granger, ``Transinpaint: Transformer-based image inpainting with context adaptation,'' in \emph{Proc. Int. Conf. Comput. Vis.}, 2023, pp. 849--858.

\bibitem{wan2021high}
Z.~Wan, J.~Zhang, D.~Chen, and J.~Liao, ``High-fidelity pluralistic image completion with transformers,'' in \emph{Proc. Int. Conf. Comput. Vis.}, 2021.

\bibitem{shamsolmoali2023vtae}
P.~Shamsolmoali, M.~Zareapoor, H.~Zhou, D.~Tao, and X.~Li, ``Vtae: Variational transformer autoencoder with manifolds learning,'' \emph{IEEE Trans. Image Process.}, 2023.

\bibitem{ho2020denoising}
J.~Ho, A.~Jain, and P.~Abbeel, ``Denoising diffusion probabilistic models,'' \emph{Proc. Advances Neural Inf. Process. Syst.}, vol.~33, pp. 6840--6851, 2020.

\bibitem{song2020denoising}
J.~Song, C.~Meng, and S.~Ermon, ``Denoising diffusion implicit models,'' \emph{Proc. Int. Conf. Learn. Represent.}, 2021.

\bibitem{wang2022zero}
Y.~Wang, J.~Yu, and J.~Zhang, ``Zero-shot image restoration using denoising diffusion null-space,'' \emph{Proc. Int. Conf. Learn. Represent.}, 2023.

\bibitem{avrahami2022blended}
O.~Avrahami, D.~Lischinski, and O.~Fried, ``Blended diffusion for text-driven editing of natural images,'' in \emph{Proc. Conf. Comput. Vis. Pattern Recognit.}, 2022, pp. 18\,208--18\,218.

\bibitem{lugmayr2022repaint}
A.~Lugmayr, M.~Danelljan, A.~Romero, F.~Yu, R.~Timofte, and L.~Van~Gool, ``Repaint: Inpainting using denoising diffusion probabilistic models,'' in \emph{Proc. Conf. Comput. Vis. Pattern Recognit.}, 2022.

\bibitem{trippe2022diffusion}
B.~L. Trippe, J.~Yim, D.~Tischer, D.~Baker, T.~Broderick, R.~Barzilay, and T.~Jaakkola, ``Diffusion probabilistic modeling of protein backbones in 3d for the motif-scaffolding problem,'' \emph{Proc. Advances Neural Inf. Process. Syst. workshop}, 2022.

\bibitem{kong2021fast}
Z.~Kong and W.~Ping, ``On fast sampling of diffusion probabilistic models,'' \emph{Proc. Int. Conf. Mach. Learn. workshop}, 2021.

\bibitem{jonnarth2022importance}
A.~Jonnarth and M.~Felsberg, ``Importance sampling cams for weakly-supervised segmentation,'' in \emph{Proc. Int. Conf. Acoust. Speech Signal Process.}, 2022, pp. 2639--2643.

\bibitem{liu2022pseudo}
L.~Liu, Y.~Ren, Z.~Lin, and Z.~Zhao, ``Pseudo numerical methods for diffusion models on manifolds,'' \emph{Proc. Int. Conf. Learn. Represent.}, 2022.

\bibitem{tong2021diffusion}
A.~Y. Tong, G.~Huguet, A.~Natik, K.~MacDonald, M.~Kuchroo, R.~Coifman, G.~Wolf, and S.~Krishnaswamy, ``Diffusion earth mover’s distance and distribution embeddings,'' in \emph{Proc. Int. Conf. Mach. Learn.}, 2021.

\bibitem{li2023dpm}
Z.~Li, S.~Li, Z.~Wang, N.~Lei, Z.~Luo, and D.~X. Gu, ``Dpm-ot: a new diffusion probabilistic model based on optimal transport,'' in \emph{Proc. Int. Conf. Comput. Vis.}, 2023, pp. 22\,624--22\,633.

\bibitem{ozair2019wasserstein}
S.~Ozair, C.~Lynch, Y.~Bengio, A.~Van~den Oord, S.~Levine, and P.~Sermanet, ``Wasserstein dependency measure for representation learning,'' \emph{Proc. Advances Neural Inf. Process. Syst.}, vol.~32, 2019.

\bibitem{chung2022diffusion}
H.~Chung, J.~Kim, M.~T. Mccann, M.~L. Klasky, and J.~C. Ye, ``Diffusion posterior sampling for general noisy inverse problems,'' \emph{Proc. Int. Conf. Learn. Represent.}, 2023.

\bibitem{chung2022improving}
H.~Chung, B.~Sim, D.~Ryu, and J.~C. Ye, ``Improving diffusion models for inverse problems using manifold constraints,'' \emph{Proc. Advances Neural Inf. Process. Syst.}, vol.~35, pp. 25\,683--25\,696, 2022.

\bibitem{zhang2023towards}
G.~Zhang, J.~Ji, Y.~Zhang, M.~Yu, T.~S. Jaakkola, and S.~Chang, ``Towards coherent image inpainting using denoising diffusion implicit models,'' \emph{Proc. Int. Conf. Mach. Learn.}, 2023.

\bibitem{lu2019brenier}
Y.~Lu, L.~Chen, A.~Saidi, and X.~Gu, ``Brenier approach for optimal transportation between a quasi-discrete measure and a discrete measure,'' 2019, pp. 204--212.

\bibitem{paty2020regularity}
F.-P. Paty, A.~d’Aspremont, and M.~Cuturi, ``Regularity as regularization: Smooth and strongly convex brenier potentials in optimal transport,'' in \emph{Proc. Int. Conf. Artif. Intell. Statis.}, 2020, pp. 1222--1232.

\bibitem{papadakis2015optimal}
N.~Papadakis, ``Optimal transport for image processing,'' Ph.D. dissertation, Universit{\'e} de Bordeaux; Habilitation thesis, 2015.

\bibitem{barnes2009patchmatch}
C.~Barnes, E.~Shechtman, A.~Finkelstein, and D.~B. Goldman, ``Patchmatch: A randomized correspondence algorithm for structural image editing,'' \emph{ACM Trans. Graph.}, vol.~28, no.~3, p.~24, 2009.

\bibitem{iizuka2017globally}
S.~Iizuka, E.~Simo-Serra, and H.~Ishikawa, ``Globally and locally consistent image completion,'' \emph{ACM Trans. Graph.}, vol.~36, no.~4, 2017.

\bibitem{yu2018generative}
J.~Yu, Z.~Lin, J.~Yang, X.~Shen, X.~Lu, and T.~S. Huang, ``Generative image inpainting with contextual attention,'' in \emph{Proc. Conf. Comput. Vis. Pattern Recognit.}, 2018, pp. 5505--5514.

\bibitem{zhang2018semantic}
H.~Zhang, Z.~Hu, C.~Luo, W.~Zuo, and M.~Wang, ``Semantic image inpainting with progressive generative networks,'' in \emph{Proc. ACM Int. Conf. Multimedia}, 2018, pp. 1939--1947.

\bibitem{liu2020rethinking}
H.~Liu, B.~Jiang, Y.~Song, W.~Huang, and C.~Yang, ``Rethinking image inpainting via a mutual encoder-decoder with feature equalizations,'' in \emph{Proc. Eur. Conf. Comput. Vis.}, 2020, pp. 725--741.

\bibitem{zeng2021cr}
Y.~Zeng, Z.~Lin, H.~Lu, and V.~M. Patel, ``Cr-fill: Generative image inpainting with auxiliary contextual reconstruction,'' in \emph{Proc. Conf. Comput. Vis. Pattern Recognit.}, 2021, pp. 14\,164--14\,173.

\bibitem{dong2022incremental}
Q.~Dong, C.~Cao, and Y.~Fu, ``Incremental transformer structure enhanced image inpainting with masking positional encoding,'' in \emph{Proc. Conf. Comput. Vis. Pattern Recognit.}, 2022, pp. 11\,358--11\,368.

\bibitem{jain2023keys}
J.~Jain, Y.~Zhou, and N.~Yu, ``Keys to better image inpainting: Structure and texture go hand in hand,'' in \emph{Winter Conf. Appl. Comput. Vis.}, 2023.

\bibitem{zhao2021large}
S.~Zhao, J.~Cui, Y.~Sheng, Y.~Dong, X.~Liang, E.~I. Chang, and Y.~Xu, ``Large scale image completion via co-modulated generative adversarial networks,'' \emph{Proc. Int. Conf. Learn. Representations}, 2021.

\bibitem{liu2018image}
G.~Liu, F.~A. Reda, K.~J. Shih, T.-C. Wang, A.~Tao, and B.~Catanzaro, ``Image inpainting for irregular holes using partial convolutions,'' in \emph{Proc. Eur. Conf. Comput. Vis.}, 2018, pp. 85--100.

\bibitem{yu2020region}
T.~Yu, Z.~Guo, X.~Jin, S.~Wu, Z.~Chen, W.~Li, Z.~Zhang, and S.~Liu, ``Region normalization for image inpainting,'' in \emph{Proc. AAAI}, 2020.

\bibitem{dosovitskiy2020image}
A.~Dosovitskiy, L.~Beyer, A.~Kolesnikov, D.~Weissenborn, X.~Zhai, T.~Unterthiner, M.~Dehghani, M.~Minderer, G.~Heigold, S.~Gelly \emph{et~al.}, ``An image is worth 16x16 words: Transformers for image recognition at scale,'' \emph{Proc. Int. Conf. Learn. Representations}, 2020.

\bibitem{yu2021diverse}
Y.~Yu, F.~Zhan, R.~Wu, J.~Pan, K.~Cui, S.~Lu, F.~Ma, X.~Xie, and C.~Miao, ``Diverse image inpainting with bidirectional and autoregressive transformers,'' in \emph{Proc. Int. Conf. Multimedia}, 2021, pp. 69--78.

\bibitem{liao2021image}
L.~Liao, J.~Xiao, Z.~Wang, C.-W. Lin, and S.~Satoh, ``Image inpainting guided by coherence priors of semantics and textures,'' in \emph{Proc. Conf. Comput. Vis. Pattern Recognit.}, 2021, pp. 6539--6548.

\bibitem{nazeri2019edgeconnect}
K.~Nazeri, E.~Ng, T.~Joseph, F.~Z. Qureshi, and M.~Ebrahimi, ``Edgeconnect: Generative image inpainting with adversarial edge learning,'' \emph{Proc. Int. Conf. Comput. Vis. Workshop}, 2019.

\bibitem{yi2020contextual}
Z.~Yi, Q.~Tang, S.~Azizi, D.~Jang, and Z.~Xu, ``Contextual residual aggregation for ultra high-resolution image inpainting,'' in \emph{Proc. Conf. Comput. Vis. Pattern Recognit.}, 2020, pp. 7508--7517.

\bibitem{bond2022unleashing}
S.~Bond-Taylor, P.~Hessey, H.~Sasaki, T.~P. Breckon, and C.~G. Willcocks, ``Unleashing transformers: Parallel token prediction with discrete absorbing diffusion for fast high-resolution image generation from vector-quantized codes,'' in \emph{Proc. Eur. Conf. Comput. Vis.}, 2022, pp. 170--188.

\bibitem{chung2022come}
H.~Chung, B.~Sim, and J.~C. Ye, ``Come-closer-diffuse-faster: Accelerating conditional diffusion models for inverse problems through stochastic contraction,'' in \emph{Proc. Conf. Comput. Vis. Pattern Recognit.}, 2022.

\bibitem{bansal2022cold}
A.~Bansal, E.~Borgnia, H.-M. Chu, J.~S. Li, H.~Kazemi, F.~Huang, M.~Goldblum, J.~Geiping, and T.~Goldstein, ``Cold diffusion: Inverting arbitrary image transforms without noise,'' \emph{arXiv preprint arXiv:2208.09392}, 2022.

\bibitem{liu2022delving}
H.~Liu, Y.~Wang, M.~Wang, and Y.~Rui, ``Delving globally into texture and structure for image inpainting,'' in \emph{Proc. ACM Int. Conf. Multimedia}, 2022, pp. 1270--1278.

\bibitem{su2022dual}
X.~Su, J.~Song, C.~Meng, and S.~Ermon, ``Dual diffusion implicit bridges for image-to-image translation,'' \emph{Proc. Int. Conf. Learn. Represent.}, 2023.

\bibitem{wu2023slotdiffusion}
Z.~Wu, J.~Hu, W.~Lu, I.~Gilitschenski, and A.~Garg, ``Slotdiffusion: Object-centric generative modeling with diffusion models,'' \emph{Proc. Advances Neural Inf. Process. Syst.}, 2023.

\bibitem{zhang2022gddim}
Q.~Zhang, M.~Tao, and Y.~Chen, ``gddim: Generalized denoising diffusion implicit models,'' \emph{Proc. Int. Conf. Learn. Represent.}, 2023.

\bibitem{song2019generative}
Y.~Song and S.~Ermon, ``Generative modeling by estimating gradients of the data distribution,'' \emph{Proc. Advances Neural Inf. Process. Syst.}, 2019.

\bibitem{kawar2022denoising}
B.~Kawar, M.~Elad, S.~Ermon, and J.~Song, ``Denoising diffusion restoration models,'' \emph{Proc. Advances Neural Inf. Process. Syst.}, 2022.

\bibitem{saharia2022palette}
C.~Saharia, W.~Chan, H.~Chang, C.~Lee, J.~Ho, T.~Salimans, D.~Fleet, and M.~Norouzi, ``Palette: Image-to-image diffusion models,'' in \emph{Proc. ACM SIGGRAPH Conf.}, 2022, pp. 1--10.

\bibitem{nichol2021glide}
A.~Nichol, P.~Dhariwal, A.~Ramesh, P.~Shyam, P.~Mishkin, B.~McGrew, I.~Sutskever, and M.~Chen, ``Glide: Towards photorealistic image generation and editing with text-guided diffusion models,'' \emph{Proc. Int. Conf. Learn. Represent.}, 2021.

\bibitem{xie2023smartbrush}
S.~Xie, Z.~Zhang, Z.~Lin, T.~Hinz, and K.~Zhang, ``Smartbrush: Text and shape guided object inpainting with diffusion model,'' in \emph{Proc. Conf. Comput. Vis. Pattern Recognit.}, 2023, pp. 22\,428--22\,437.

\bibitem{rombach2022high}
R.~Rombach, A.~Blattmann, D.~Lorenz, P.~Esser, and B.~Ommer, ``High-resolution image synthesis with latent diffusion models,'' in \emph{Proc. Conf. Comput. Vis. Pattern Recognit.}, 2022.

\bibitem{song2022pseudoinverse}
J.~Song, A.~Vahdat, M.~Mardani, and J.~Kautz, ``Pseudoinverse-guided diffusion models for inverse problems,'' in \emph{Proc. Int. Conf. Learn. Represent.}, 2023.

\bibitem{liu20232}
G.-H. Liu, A.~Vahdat, D.-A. Huang, E.~A. Theodorou, W.~Nie, and A.~Anandkumar, ``I$^{2}$sb: Image-to-image schrodinger bridge,'' \emph{Proc. Int. Conf. Mach. Learn.}, 2023.

\bibitem{chung2024direct}
H.~Chung, J.~Kim, and J.~C. Ye, ``Direct diffusion bridge using data consistency for inverse problems,'' \emph{Proc. Advances Neural Inf. Process. Syst.}, vol.~36, 2024.

\bibitem{zhang2023adding}
L.~Zhang, A.~Rao, and M.~Agrawala, ``Adding conditional control to text-to-image diffusion models,'' in \emph{Proc. Int. Conf. Comput. Vis.}, 2023, pp. 3836--3847.

\bibitem{zhuang2025task}
J.~Zhuang, Y.~Zeng, W.~Liu, C.~Yuan, and K.~Chen, ``A task is worth one word: Learning with task prompts for high-quality versatile image inpainting,'' in \emph{Proc. Eur. Conf. Comput. Vis.}, 2025, pp. 195--211.

\bibitem{pokle2022deep}
A.~Pokle, Z.~Geng, and J.~Z. Kolter, ``Deep equilibrium approaches to diffusion models,'' \emph{Proc. Advances Neural Inf. Process. Syst.}, vol.~35, pp. 37\,975--37\,990, 2022.

\bibitem{dhariwal2021diffusion}
P.~Dhariwal and A.~Nichol, ``Diffusion models beat gans on image synthesis,'' \emph{Proc. Advances Neural Inf. Process. Syst.}, 2021.

\bibitem{panaretos2019statistical}
V.~M. Panaretos and Y.~Zemel, ``Statistical aspects of wasserstein distances,'' \emph{Annu. Rev. Stat. Appl.}, vol.~6, no.~1, pp. 405--431, 2019.

\bibitem{mokrov2021large}
P.~Mokrov, A.~Korotin, L.~Li, A.~Genevay, J.~M. Solomon, and E.~Burnaev, ``Large-scale wasserstein gradient flows,'' \emph{Proc. Advances Neural Inf. Process. Syst.}, vol.~34, pp. 15\,243--15\,256, 2021.

\bibitem{fan2021variational}
J.~Fan, Q.~Zhang, A.~Taghvaei, and Y.~Chen, ``Variational wasserstein gradient flow,'' \emph{Proc. Int. Conf. Mach. Learn.}, 2022.

\bibitem{liu2015deep}
Z.~Liu, P.~Luo, X.~Wang, and X.~Tang, ``Deep learning face attributes in the wild,'' in \emph{Proc. Int. Conf. Comput. Vis.}, 2015, pp. 3730--3738.

\bibitem{zhou2017places}
B.~Zhou, A.~Lapedriza, A.~Khosla, A.~Oliva, and A.~Torralba, ``Places: A 10 million image database for scene recognition,'' \emph{IEEE Trans. Pattern Anal. Mach. Intell.}, vol.~40, no.~6, pp. 1452--1464, 2017.

\bibitem{russakovsky2015imagenet}
O.~Russakovsky, J.~Deng, H.~Su, J.~Krause, S.~Satheesh, S.~Ma, Z.~Huang, A.~Karpathy, A.~Khosla, M.~Bernstein \emph{et~al.}, ``Imagenet large scale visual recognition challenge,'' \emph{Int. J. Comput. Vis.}, vol. 115, no.~3, 2015.

\bibitem{romero2022ntire}
A.~Romero, A.~Castillo, J.~Abril-Nova, R.~Timofte, R.~Das, S.~Hira, Z.~Pan, M.~Zhang, B.~Li, D.~He \emph{et~al.}, ``Ntire 2022 image inpainting challenge: Report,'' in \emph{Proc. Conf. Comput. Vis. Pattern Recognit.}, 2022.

\end{thebibliography}
}}
\begin{IEEEbiography}
[{\includegraphics[width=1in,height=1in,clip,keepaspectratio]{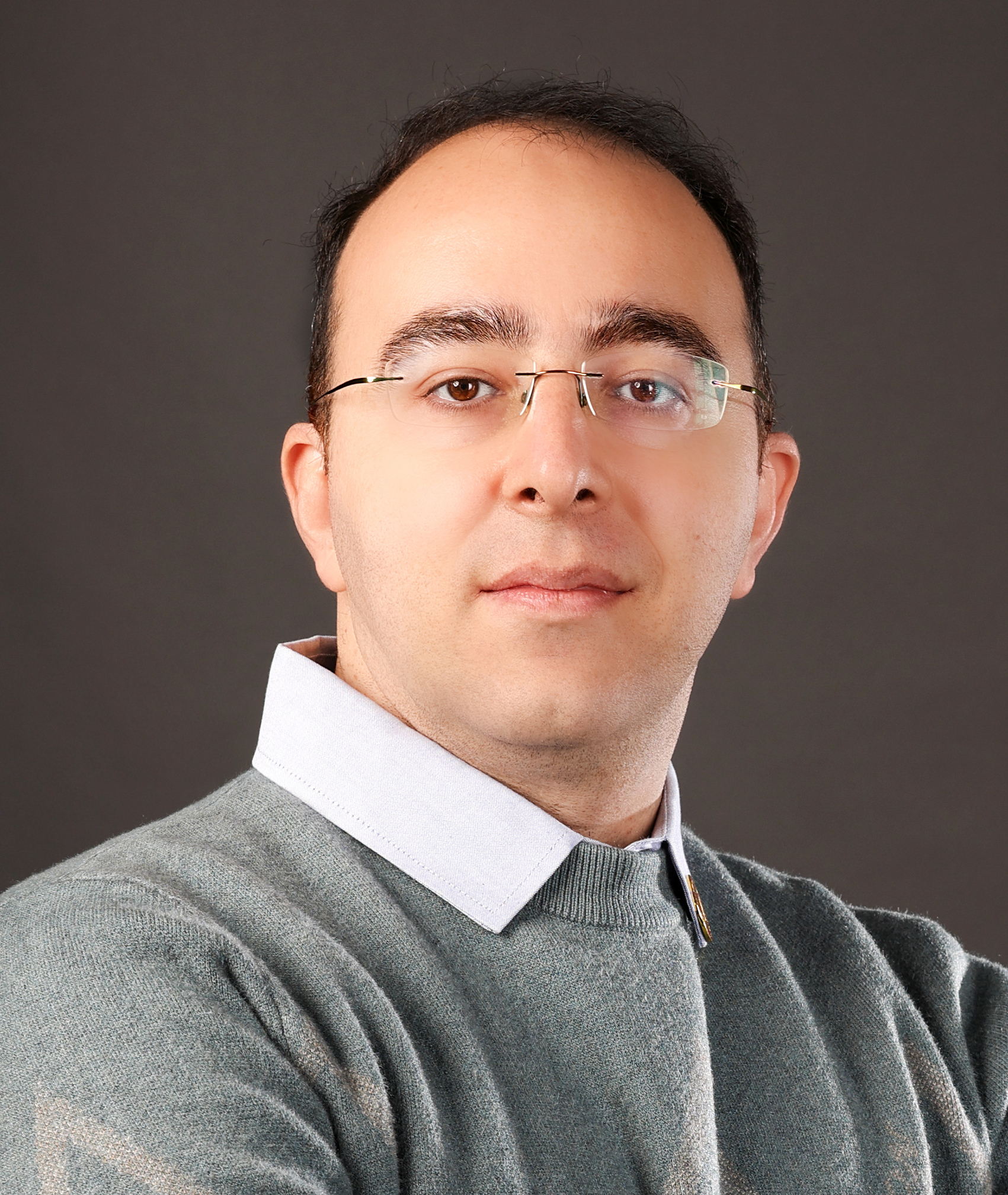}}]
{Pourya Shamsolmoali} (Senior Member, IEEE) received the PhD degree in computer science from Shanghai Jiao Tong University. He has been a visiting researcher with Linköping University, INRIA-France, and ETS-Montreal. He is currently a lecturer with the University of York. His main research focuses on machine
learning, computer vision, and image processing.
\end{IEEEbiography}
\begin{IEEEbiography}
[{\includegraphics[width=1in,height=1in,clip,keepaspectratio]{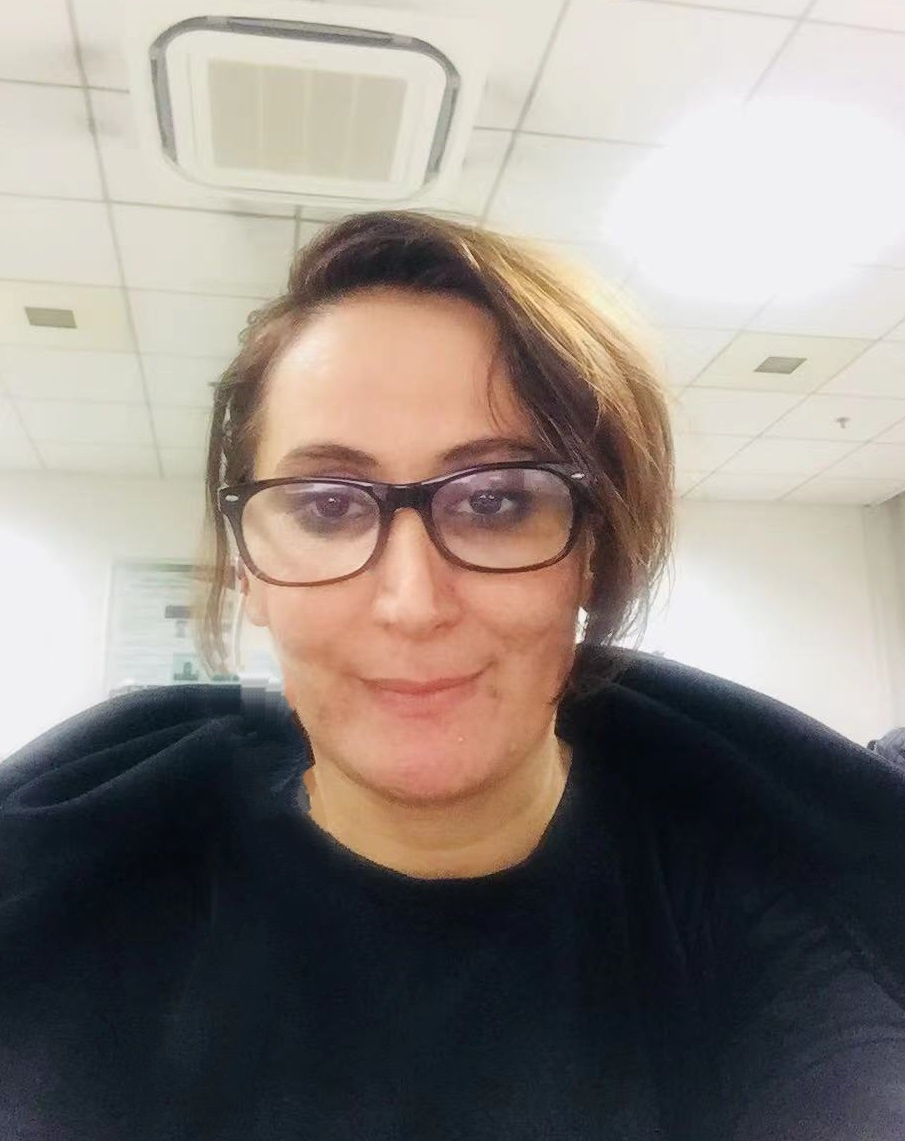}}]
{Masoumeh Zareapoor} (Member, IEEE) received the PhD degree in computer science from Jamia University, India. She is an Associate Researcher at Shanghai Jiao Tong University and was a visiting researcher at Northwestern Polytechnical University, Xi'an. Additionally, she was a research engineer at Huawei, Shanghai. Her main research focuses on machine learning and computer vision.
\end{IEEEbiography}
\begin{IEEEbiography}
[{\includegraphics[width=1in,height=1in,clip,keepaspectratio]{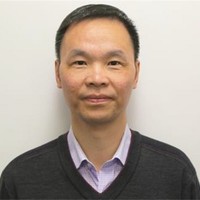}}]
{Huiyu Zhou} received the PhD degree in computer vision from Heriot-Watt University, Edinburgh, U.K. He currently is a full professor with the School of Computing and Mathematical Sciences, University of Leicester, U.K. He has published widely in the field. His research work has been or is being supported by U.K. EPSRC, ESRC, AHRC, MRC, EU, Royal Society, Leverhulme Trust, Puffin Trust, Invest NI and industry.
\end{IEEEbiography}
\begin{IEEEbiography}
[{\includegraphics[width=1in,height=1in,clip,keepaspectratio]{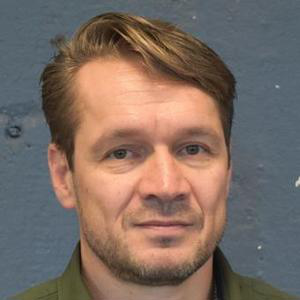}}]
{Michael Felsberg} (Senior Member, IEEE) received the PhD degree from Kiel University, Germany, in 2002, and the docent degree from Linköping University, in 2005. He is a full professor with Linköping University, Sweden, since 2008. He received the DAGM Olympus award in 2005 and is fellow of the IAPR and ELLIS.
His research interests include video object and instance segmentation,
classification, segmentation, and registration of point clouds, as well
as efficient machine learning techniques for incremental, few-shot, and
long-tailed settings.
\end{IEEEbiography}
\begin{IEEEbiography}
[{\includegraphics[width=1in,height=1in,clip,keepaspectratio]{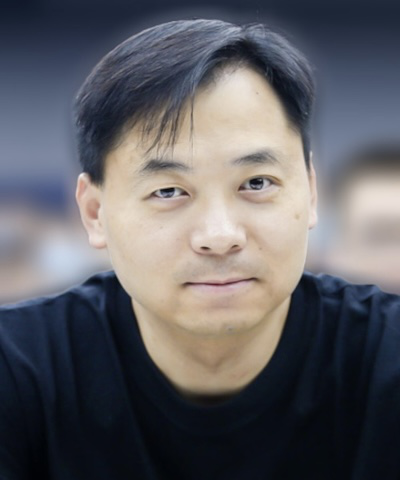}}]
{Dacheng Tao} (IEEE Fellow) is currently a Distinguished University Professor in the College of Computing \& Data Science at Nanyang Technological University. He mainly applies statistics and mathematics to artificial intelligence, data science, and his research is detailed in one monograph and over 200 publications in prestigious journals and proceedings at leading conferences, with best paper awards, best student paper awards, and test-of-time awards. His publications have been cited over 112K times and he has an h-index of 160+ in Google Scholar. He received the 2015 and 2020 Australian Eureka Prize, the 2018 IEEE ICDM Research Contributions Award, the 2019 Diploma of The Polish Neural Network Society, and the 2021 IEEE Computer Society McCluskey Technical Achievement Award. He is a Fellow of the Australian Academy of Science, AAAS, ACM and IEEE. 
\end{IEEEbiography}
\begin{IEEEbiography}
[{}]
{Xuelong Li} (Fellow, IEEE) is the CTO and Chief Scientist of China Telecom, where he founded the Institute of Artificial Intelligence (TeleAI) of China Telecom.
\end{IEEEbiography}

\end{document}